\documentclass[lettersize,journal]{IEEEtran}

\newtheorem{theorem}{Theorem}
\newtheorem{corollary}{Corollary}
\newtheorem{lemma}{Lemma}   

\newtheorem{assumption}{Assumption}
\newtheorem{remark}{Remark}

\usepackage[linesnumbered,ruled,boxed]{algorithm2e}
\usepackage{amsmath,amsfonts}
\usepackage{bbm}
\usepackage{algorithmic}
\usepackage{array}
\usepackage[caption=false,font=normalsize,labelfont=sf,textfont=sf]{subfig}
\usepackage{textcomp}
\usepackage{stfloats}
\usepackage{adjustbox}
 \usepackage{multirow}
 \usepackage{booktabs}       
\usepackage{url}
\usepackage{verbatim}
\usepackage{graphicx}
\usepackage{caption} 
\usepackage{color}

\def\BibTeX{{\rm B\kern-.05em{\sc i\kern-.025em b}\kern-.08em
    T\kern-.1667em\lower.7ex\hbox{E}\kern-.125emX}}
\usepackage{balance}
\begin{document}
\title{ Online Optimization  for Learning to Communicate over Time-Correlated  Channels}
    \author{Zheshun Wu, Junfan Li, Zenglin Xu, \IEEEmembership{Senior Member, IEEE}, Sumei Sun, \IEEEmembership{Fellow, IEEE}, and Jie Liu, \IEEEmembership{Fellow, IEEE}
\thanks{Copyright (c) 2015 IEEE. Personal use of this material is permitted. However, permission to use this material for any other purposes must be obtained from the IEEE by sending a request to pubs-permissions@ieee.org.}
\thanks{This work was supported in part by National Natural Science Foundation
of China (No. 62350710797). (\emph{Corresponding author: Jie Liu; Zenglin Xu.})}
\thanks{Zheshun Wu, Junfan Li and Jie Liu are with the School of Computer Science and Technology, Harbin Institute of Technology Shenzhen, Shenzhen 518055, China (e-mail:
wuzhsh23@gmail.com; lijunfan@hit.edu.cn; jieliu@hit.edu.cn).}
\thanks{Zenglin Xu is with the Fudan University, and also with the Shanghai Academy of Artificial Intelligence for Science  (e-mail: zenglin@gmail.com).}
\thanks{Sumei Sun is with the Institute for Infocomm Research, Agency for Science, Technology and Research, Singapore 138632 (e-mail: sunsm@i2r.a-star.edu.sg).}}


\maketitle

\begin{abstract}
 Machine learning techniques have garnered great interest in designing communication systems owing to their capacity in tackling with channel uncertainty. To provide theoretical guarantees for learning-based communication systems, some recent works analyze generalization bounds for devised methods based on the assumption of Independently and Identically Distributed (I.I.D.) channels, a condition rarely met in practical scenarios. In this paper, we drop the I.I.D. channel assumption and study an online optimization problem of learning to communicate over time-correlated channels. To address this issue, we further focus on two specific tasks: optimizing channel decoders for time-correlated fading channels and selecting optimal codebooks for time-correlated additive noise channels. For utilizing temporal dependence of considered channels to better learn communication systems, we develop two online optimization algorithms based on the optimistic online mirror descent framework. Furthermore, we provide theoretical guarantees for proposed algorithms via deriving sub-linear regret bound on the expected error probability of learned systems. Extensive simulation experiments have been conducted to validate that our presented approaches can leverage the channel correlation to achieve a lower average symbol error rate compared to  baseline methods, consistent with our theoretical findings.
\end{abstract}

\begin{IEEEkeywords}
Time-correlated channels, decoder learning, codebook selection, online optimization theory, online convex optimization, multi-armed bandit, error probability analysis.
\end{IEEEkeywords}

\section{Introduction}
\IEEEPARstart{T}{he} widespread adoption of machine learning techniques in developing communication systems based on real-world data has sparked broad interest in recent years~\cite{8542764}. Machine learning algorithms have shown to be effective tools for various tasks such as channel estimation~\cite{bai2019deep},  coding~\cite{bourtsoulatze2019deep}, decoding~\cite{DBLP:conf/nips/NachmaniW19}, and other physical layer applications~\cite{DBLP:journals/tccn/OSheaH17}.  Learning-based communication systems have showed impressive performance in their capacity to generalize effectively to unknown channels~\cite{DBLP:journals/wc/HuangGGYZSA20}.
Most existing studies utilizing machine learning approaches to design communication systems lack theoretical justification for their proposed methods, and typically regard learned communication systems as black boxes~\cite{ DBLP:journals/tccn/OSheaH17,DBLP:journals/wc/HuangGGYZSA20}. Recently, a few works are aiming to conduct theoretical analyses for learning-based communication systems~\cite{weinberger2021generalization,DBLP:conf/isit/BernardoZE23,liu2024pac}. 
Specifically, these studies leverage statistical learning theory~\cite{shalev2014understanding,vapnik2013nature} to derive generalization bounds based on the Independently and Identically Distributed (I.I.D.) channel assumption.  Building upon this assumption, the generalization capacity of learning-based communication systems can be assured. This is because these systems can be trained using sufficient data during the offline stage and demonstrate commendable performance when evaluated with data sampled from the same distribution. 
   
However, meeting the assumption of I.I.D. channels in reality is generally challenging, given that practical communication scenarios often involve time-varying channels with dynamic statistical properties. Among time-varying channels, a prominent example is the time-correlated channel~\cite{DBLP:journals/inffus/Caballero-Aguila20}.  In real-world communication processes, there exist numerous instances of time-correlated channels. For example, user mobility usually leads to time-correlated fading channels in mobile communication~\cite{DBLP:journals/tmc/YaoBS23,DBLP:journals/tsmc/TanSS22}. Specifically, in vehicular ad-hoc network scenarios (VANET),  wireless links between moving vehicles and roadside units (RSUs) are significantly affected by vehicles' mobility~\cite{DBLP:journals/access/ZhuYCTFWL18}. As a consequence of moderate user mobility, channel fading gains become interdependent between consecutive time slots, resulting in the well-known Markov fading channels~\cite{DBLP:journals/isci/WangD24,DBLP:journals/tit/LarranagaADP18}. Notably, the nonlinear behavior of hardware components transforms additive white noise in the receiver into colored noise, thereby introducing temporal correlation~\cite{DBLP:journals/tcom/ChavaliS13}. Besides, time-correlated channel noise can  arise due to narrowband filtering of uncorrelated impulsive noise by the RF front-end~\cite{542437}. In these scenarios,  establishing generalization bounds for learning-based communication systems over time-correlated channels is challenging. This is because classic tools like concentration inequalities from statistical learning theory~\cite{mohri2018foundations}, which heavily depend on the strong I.I.D. channel assumption, are not applicable in these scenarios.


   \begin{figure*}[ht]
\captionsetup[subfloat]{font=footnotesize}	
\centering
   	\subfloat[learn decoder via training sequence]{\includegraphics[width = 0.45\textwidth]{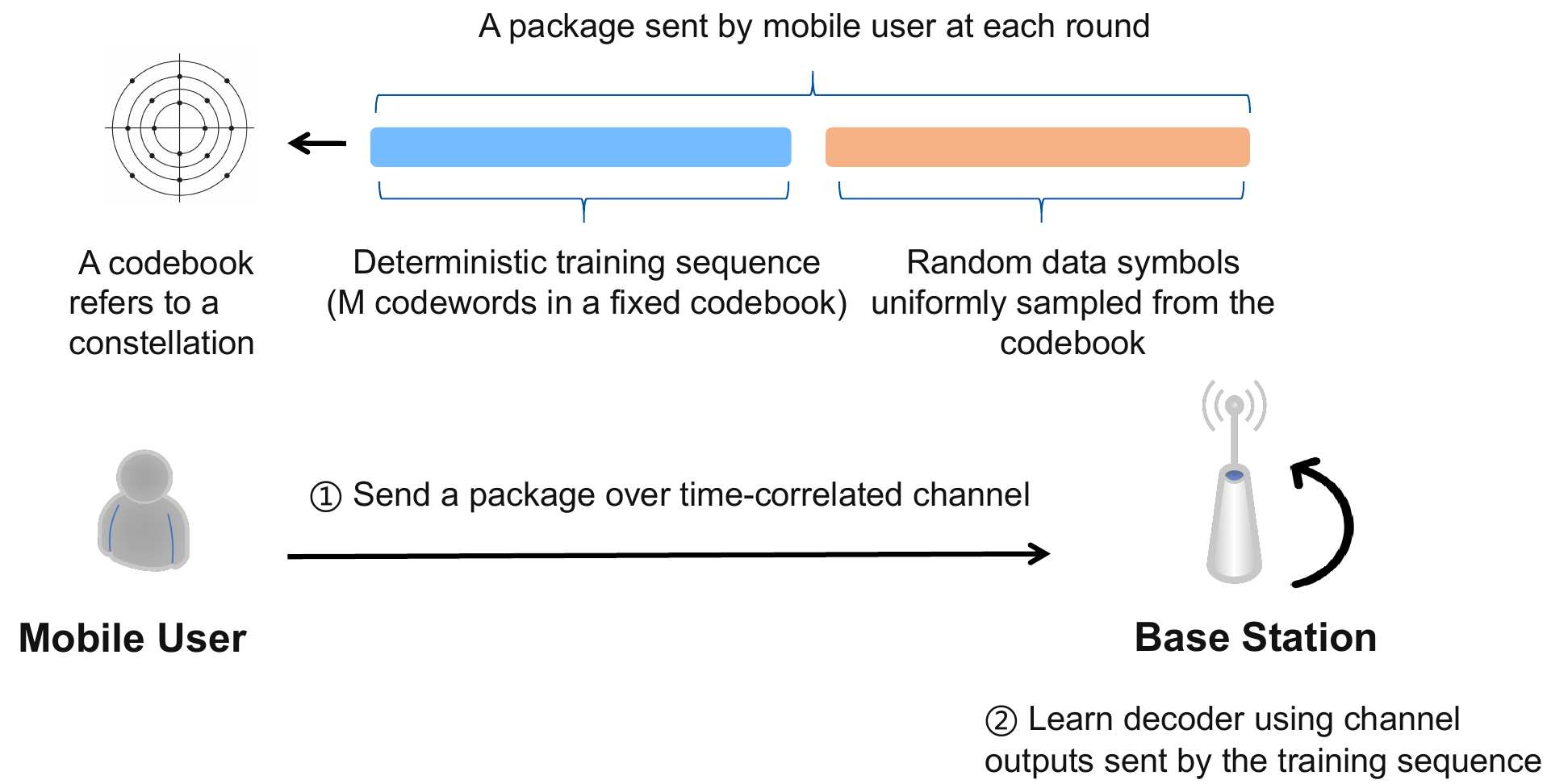}}
    \hspace{0.2in}
    \subfloat[learn codebook via training sequence]
    {\includegraphics[width = 0.45\textwidth]{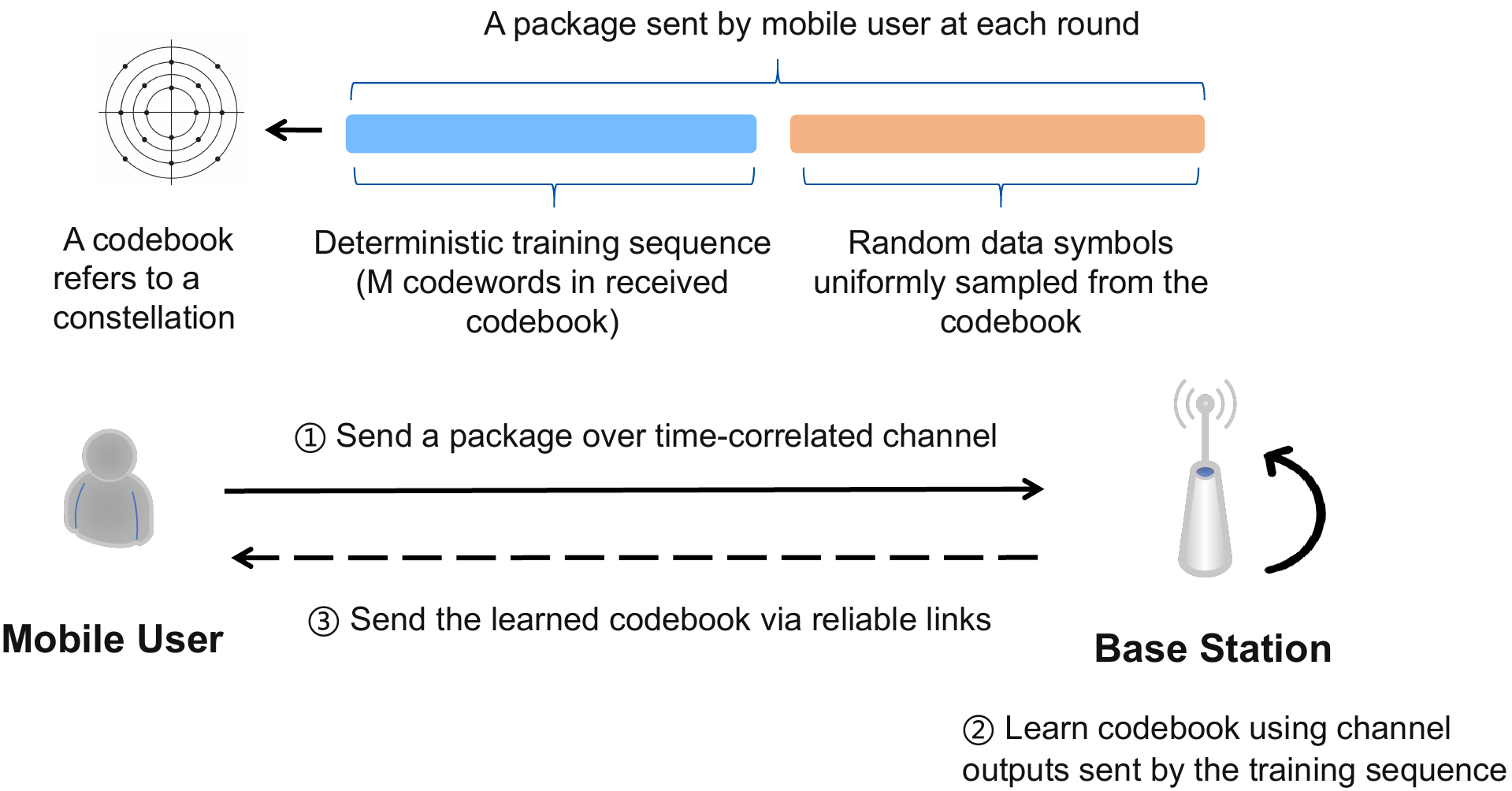}}
\caption{The procedure of  learning decoder and codebook via online optimization. }
\label{fig:system}
\end{figure*}

\begin{figure*}[ht]
    \centering
    \includegraphics[width=5.in]{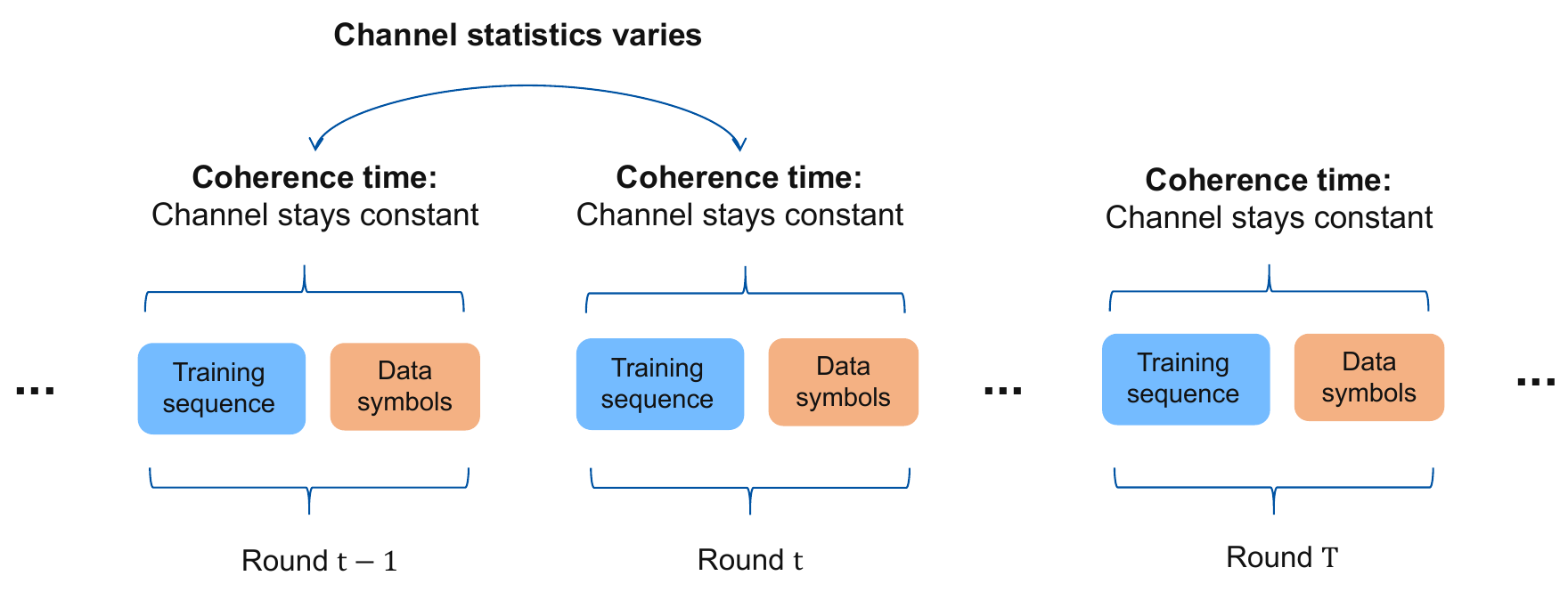}
    \caption{Transmission process of uplink communication from mobile user to base station.}
    \label{fig:coherence}
    \vspace{-0.2in}
\end{figure*}

In this paper, we will establish online optimization algorithms~\cite{shalev2012online,hazan2016introduction} for learning communication systems over time-correlated channels with solid theoretical guarantees. Specifically, we consider optimizing the uplink communication process between a mobile user equipment acting as the transmitter and a base station serving as a receiver.   One potential application of proposed algorithms is to enhance the performance of distributed learning conducted by multiple mobile devices within wireless networks~\cite{DBLP:journals/spl/WuXYL24,DBLP:journals/tvt/WuXZLL24}. Notably, our framework directly supports learning the downlink component of communication systems by enabling users to execute online optimization algorithms locally. In each communication round, we consider that the transmitter sends a package to the receiver. This package   is composed of two parts: the first part is a deterministic training sequence known to both the transmitter and receiver, and the second part consists of data symbols that carry the actual information intended for transmission. These data symbols are random and unknown to the receiver. We highlight that this training sequence serves a similar role of pilot signals used in channel estimation in mobile communication, which is essential for calculating the loss required for conducting online optimization. The outlined online optimization procedure is depicted in Fig.~\ref{fig:system}.  Following the assumption widely adopted in the study on using online learning to optimize communication systems~\cite{DBLP:journals/tcom/ZhangHZY20,wei2022fast}, we  presume that the channel remains constant during the transmission of both the training sequence and the data symbols within a single communication round. While the channel statistics varies across different rounds. Consequently,  channel decoders and codebooks learned by training sequences can be utilized to improve the transmission performance of data symbols within the same round. The transmission process of uplink communication is shown in Fig.~\ref{fig:coherence}.

For the time-correlated channels studied in this paper, to be specific, we focus on two types of channels: 1) \emph{time-correlated fading channel} with additive white Gaussian noise, where the fading distribution is unknown and time-correlated; and 2) \emph{time-correlated additive noise channel} with an unknown and time-correlated noise distribution. For the time-correlated fading channel, the transceiver is equipped with a fixed codebook (constellation) and is tasked with optimizing the channel decoder using collected channel input-output pairs data.  The optimized decoder processes channel outputs to compensate for fading-induced distortions—similar to channel equalization—thus enhancing symbol detection and overall transmission performance.
Besides, based on the proposed convex surrogate loss, we can regard such online channel decoder learning problem as an online convex optimization problem. In the context of time-correlated additive noise channels with deterministic identity channel gains, the primary task of transceivers shifts to optimizing codebook selection to mitigate the effects of channel noise and enhance overall transmission performance. Similar to~\cite{weinberger2021generalization}, the codebook considered in this paper essentially refers to a modulation scheme, i.e., a constellation.
We model this online codebook learning problem as a multi-armed bandit problem: the transceiver selects optimal codebooks from a predefined super-codebook using empirical data collected in real-time. 


To harness the temporal dependence of considered channels for addressing these two tasks, we separately propose an algorithm based on the optimistic Online Mirror Descent (OMD) framework within online optimization \cite{DBLP:journals/jmlr/ChiangYLMLJZ12,DBLP:conf/colt/WeiL18}.
The key advantage of proposed algorithms lies in their ability to exploit 
the inherent distribution dependency of benign environments to improve the performance of learned decisions~\cite{DBLP:conf/colt/RakhlinS13}. Then we summarize the main contributions of this work as follows,
\begin{itemize}
    \item 
            For the first time, we formulate the task of learning channel decoder or codebook for communication over time-correlated channels as an online optimization problem without relying on the I.I.D. channel assumption, which is more in line with real scenarios.
    \item We devise various algorithms to tackle the online optimization problem of learning communication systems over time-correlated channels using the optimistic OMD framework. Furthermore, we offer theoretical guarantee for our proposed algorithms.
    \item To further support our theoretical framework, we perform simulation experiments to validate the efficacy of proposed methods. Empirical results confirm that our approaches utilize channel correlation to surpass baseline methods,  matched with our theoretical discoveries.
\end{itemize}

The remainder of the paper is structured as follows: In Section II, we present a review of prior works on learning-based communication systems and online optimization theory. Section III introduces two distinct time-correlated channels and outlines the online optimization problem addressed in this study.  Section IV and V detail the development of online optimization algorithms for learning channel decoder and codebook under these two time-correlated channels respectively. Section VI encompasses numerical simulations, while Section VII concludes the paper.

\section{Related works}\label{sec:related}

In this section, we begin by presenting recent studies that have utilized machine learning techniques in the design of communication systems. Next, we provide an overview of research in online optimization, with emphasis on the application of optimistic OMD on predictable benign environments.

As previously mentioned, the remarkable success of machine learning algorithms has spurred interest in their application to optimize communication systems~\cite{DBLP:journals/tccn/OSheaH17}. Some efforts focus on replacing conventional components of communication systems with modules that are learned from empirical data~\cite{DBLP:journals/tccn/CaciularuB20}. Notable examples include~\cite{10423294} for channel estimation and equalization, and~\cite{8919799} for channel coding and decoding. Another approach enhances traditional algorithms by integrating deep neural networks~\cite{DBLP:journals/jstsp/NachmaniMLGBB18}.  For example, in~\cite{DBLP:journals/twc/ShlezingerFEG20}, DNNs are applied to enhance the Viterbi algorithm, and in~\cite{8259241}, they are used for improving the performance of belief propagation. On the other hand, some recent works leverage statistical learning theory for establishing generalization bounds of learned communication systems. For instance,~\cite{weinberger2021generalization} utilize channel input-output pair data to optimize channel decoders and constellations, and derive generalization bounds for learned communication schemes using Rademacher complexity.~\cite{DBLP:conf/isit/BernardoZE23,liu2024pac} employ the probably approximately correct (PAC) learning framework to provide theoretical guarantees for learned communication systems under discrete memoryless channels. Besides, the authors in~\cite{adiga2024generalization} derive generalization bounds for their proposed neural belief
propagation decoders.~\cite{DBLP:journals/tit/TsvieliW23} leverage the margin theory to provide theoretical guarantees for learning-based deocders. Nonetheless, previous works have predominantly conducted a theoretical analysis based on the I.I.D. channel assumption, which is not applicable to with most practical communication scenarios.

Online optimization can effectively model numerous online machine learning problems,
encompassing online convex optimization~\cite{shalev2012online}, prediction with expert advice~\cite{cesa2006prediction}, multi-armed bandit~\cite{10772999,DBLP:journals/corr/abs-2404-09494}
and more. This framework deals with a sequential decision problem, where a learner repeatedly takes actions within a feasible set and encounters a potentially adversarial loss function from the environment~\cite{shalev2012online,hazan2016introduction}.
During this decision-making process, the learner endeavors to devise an algorithm that minimizes regret, defined as the disparity between the total loss incurred by the learner and that of the best decision in hindsight.
Departing from statistical learning theory, online optimization theory 
can furnish theoretical guarantees for algorithms without relying on the I.I.D. assumption. Recently, it has been noted that practical environments often exhibit predictable patterns, which the learner can leverage to achieve reduced regret~\cite{DBLP:journals/jmlr/ChiangYLMLJZ12,DBLP:conf/colt/RakhlinS13}. 
To pursue this objective,
an optimistic Online Mirror Descent (OMD) framework has been proposed to harness environmental correlations 
for improving the performance of learned decisions~\cite{DBLP:journals/jmlr/ChiangYLMLJZ12,DBLP:conf/colt/RakhlinS13}. Furthermore, this framework is utilized to solve the semi-bandit problem in predictable environments~\cite{DBLP:conf/colt/WeiL18} and the authors in~\cite{DBLP:conf/icml/ChenT0023} leverage it to address the stochastically extended adversarial online convex optimization problem.

\section{Problem Formulation}\label{sec:problem}
\subsection{Notation Conventions}
We use standard notation or define it before its first
use, and here only focus on main conventions. The Euclidean norm for a vector $\mathbf{v}\in \mathbb{R}^d$ is denoted by $\Vert \mathbf{v}\Vert_2$. The Frobenius norm and spectral norm for a matrix $\mathbf{A}\in\mathbb{R}^{d\times d}$ are denoted by $\Vert \mathbf{A}\Vert_F:=\sqrt{tr(\mathbf{A}^T\mathbf{A})}$ and $\Vert \mathbf{A}\Vert_2:=\sup\{\Vert \mathbf{A}\mathbf{v}\Vert_2:\Vert \mathbf{v}\Vert_2=1\}$ respectively.  We define the inner product of two matrices $\mathbf{A}\in\mathbb{R}^{d\times d}$ and $\mathbf{B}\in\mathbb{R}^{d\times d}$ as $\langle \mathbf{A},\mathbf{B}\rangle=tr(\mathbf{A}^T\mathbf{B})$. For $n\in \mathbb{N}_{+}$, the set $\{1,2,...,n\}$ is denoted by $[n]$. The cardinality of
a finite set $\mathcal{X}$ is denoted by $|\mathcal{X}|$. For a set $\mathcal{X}$, $\mbox{conv}(\mathcal{X})$ means the convex hull of $\mathcal{X}$. The indicator  of an event $\mathcal{A}$ is denoted by $\mathbbm{1}\{\mathcal{A}\}$. We denote $\max\{r,0\}$ as $[r]_{+}$, where $r\in\mathbb{R}$. We use $\mathcal{O}(\cdot)$ to hide numerical constants in our upper bound and use $\Tilde{\mathcal{O}}(\cdot)$ to additionally hide logarithmic factors.

\subsection{Time-varying Channel Models}
In this paper, we assume that the transmitter sends packages to the receiver over multiple rounds, with the total number of rounds denoted as $T$. We consider the problem of communication over time-varying channels defined below:
\begin{equation}\label{eq:channel}
    Y_t=f_t(X_t)+Z_t,
\end{equation}
where $X_t\in\mathbb{R}^d$ denoting the random data symbol, is a codeword chosen from a codebook $C_t=\{\mathbf{x}_t^j\}_{j\in [M]}$ with a uniform probability. $Y_t \in \mathbb{R}^d$ is the corresponding channel output in $t$-th round. $f_t:\mathbb{R}^d \rightarrow \mathbb{R}^d$ is a channel transformation, and $Z_t\in\mathbb{R}^d$ is a channel noise statistically independent of the input $X$ and the transformation $f_t$. In this paper, we assume that the codebook $C_t\in \mathcal{C}\subset(\mathbb{R}^d)^M$ and $\mathcal{C}$ is with the  power constraint, i.e., $\mathcal{C}=\{C:  \Vert \mathbf{x}^j\Vert_2 \leq \gamma_X,\forall j\in[M]\}$.  

The statistical properties of channel transformation $f_t$ and noise $Z_t$ across different rounds are assumed to be different in this work. In other words,  we consider time-varying channels and thus the distributions of $f_t$ and $Z_t$ can vary at each round during the communication process.

In this paper, we focus on two specific cases of this time-varying channel model, both of which hold high relevance in realistic  scenarios. The first case  involves a time-correlated fading channel, assuming that $f_t$ is a time-correlated linear transformation while the channel noise $Z_t$
is an I.I.D. Gaussian noise. The second case pertains to a time-correlated additive noise channel, where $f_t$ is an identity transformation and  $Z_t$ is a time-correlated additive channel noise.

\subsubsection{Time-correlated Fading Channels}
In the first considered channel model, we fix the codebook $C_t=C=\{\mathbf{x}^j\}_{j\in[M]},\forall t \in [T]$ for the transceiver at each round. The channel transformation is assumed to be the linear transformation and it is thus similar to the common fading channel in wireless communications~\cite{goldsmith2005wireless}. Hence, Eq.~\eqref{eq:channel} becomes
\begin{equation}
    Y_t=\mathbf{H}_tX_t+W_t,
\end{equation}
where  $\mathbf{H}_t \in \mathbb{R}^{d\times d}$ is the time-correlated channel gain of this fading channel. The distribution of $\mathbf{H}_t$ is denoted by $F_{\mathbf{H}_t}$, which is different at each round $t$. Notice that we do not assume any specific distribution for $\mathbf{H}_t$, and $\{F_{\mathbf{H}_t}\}_{t \in [T]}$ is entirely unknown to the transceiver. $W_t\in\mathbb{R}^d$ is an I.I.D. additive white Gaussian
noise with zero mean and its variance is denoted by $\sigma_W^2$. The distribution of $W_t$ is denoted as $F_{W}$, and $W_t$ is independent of the input and the channel gain.

Moreover, we consider the time correlation of the channel gain $\{\mathbf{H}_t\}_{t\in [T]}$, i.e., the channel gain $\mathbf{H}_t$ depends on $\{\mathbf{H}_\tau\}_{\tau \in [t-1]}$ from previous rounds. For example, we can consider a class of first-order Markov fading channel: $\mathbf{H}_{t+1}=\sqrt{1-\mu_t}\mathbf{H}_{t}+\sqrt{\mu_t}\boldsymbol{\mathcal{E}}_{t}$, where $\boldsymbol{\mathcal{E}}_{t} \in \mathbb{R}^{d\times d}$ is a random matrix and $\mu_t \in [0,1]$ is a factor that affects the temporal dependence of channels. This channel model is associated with practical communication scenarios involving user equipments with low mobility~\cite{DBLP:journals/tmc/YaoBS23}.

For this channel, we assume that the decoding rule is chosen from  the class of nearest neighbor decoder with a linear kernel operating on the channel output, i.e., given a channel output $\mathbf{y}_t$, the index of the decoded codeword is selected as:
\begin{equation}
    \begin{aligned}
        j (\mathbf{y}_t)\in \mathop{\arg \min}\limits_{j' \in[M]}\Vert \mathbf{x}^{j'}-\mathbf{G}_t\mathbf{y}_t \Vert_2,
    \end{aligned}
\end{equation}
where  $\mathbf{y}_t=\mathbf{H}_tX_t+W_t$ and $\mathbf{G}_t\in \mathbb{R}^{d\times d}$ is a linear kernel operating on the channel output.

Now we explain this choice of decoding rule. Initially, the selection of the nearest neighbor decoder is based on its optimality under the condition that $\mathbf{G}_t=\mathbf{H}^{-1}_t$ and  $W_t$ is Gaussian~\cite{goldsmith2005wireless}, 
but we do not assume that the channel satisfies this condition here~\cite{DBLP:journals/tit/TsvieliW23}. Furthermore, the use of the linear kernel $\mathbf{G}_t$ on $\mathbf{y}_t$ shares similarities with the channel equalizer widely employed in communication systems~\cite{goldsmith2005wireless}, aiming to approximate the channel gain $\mathbf{H}_t$ and minimize its influence on decoding. For simplicity, we refer to the linear kernel $\mathbf{G}_t$ as the channel decoder in the subsequent discussions.

As shown in Fig~\ref{fig:system} (a), in each round, the transmitter sends both a deterministic training sequence including $M$ distinct codewords from $C$ and random data symbols to the receiver. We first define the $t$-th round expected error probability of learned $\mathbf{G}_t$ for $M$ codewords in the training sequence as:
\begin{equation}\label{eq:fadingprob}
\begin{aligned}
      & \mathbb{P}_t(\mathbf{G}_t):=\frac{1}{M}\sum_{j=1}^{M} P\Big\{\Vert \mathbf{x}^{j_t}-\mathbf{G}_t\mathbf{y}_t^j\Vert_2^2<\Vert \mathbf{x}^j-\mathbf{G}_t\mathbf{y}_t^j\Vert_2^2\Big\},\\
\end{aligned}
\end{equation}
where $j_t:=\arg\min_{j'\in [M]\setminus j}\Vert \mathbf{x}^{j'}-\mathbf{G}_t\mathbf{y}_t^{j}\Vert_{2}^2$ and $\mathbf{y}_t^j=\mathbf{H}_t\mathbf{x}_t^j+W_t, \forall j \in [M]$.

In this work, following the coherence time assumption considered in~\cite{wei2022fast}, we also consider that the training sequence and data symbols experience the same channel gain during a communication round. Consequently, the decoder learned by the training sequence can enhance the transmission performance of data symbols. This assumption is reasonable, as it aligns with the common practice of using similar assumptions for pilot signals in mobile communication~\cite{DBLP:journals/tcom/ZhangHZY20,9425522}. Besides, considering that data symbols are uniformly sampled from the codebook, we notice that $\mathbb{E}_{X_t}[Pr\{\hat{X}_t\neq X_t|X_t;\mathbf{G}_t\}]=\mathbb{P}_t(\mathbf{G}_t)$, where $\hat{X}_t$ represents the prediction of  the channel output transmitted by data symbol $X_t$. Based on it,  the empirical performance of $\mathbf{G}_t$  on the training sequence can be utilized to execute online optimization for identifying the optimal channel decoder on data symbols. The online optimization goal related to this motivation is to leverage channel output samples  $\{\mathbf{y}_t^j\}_{j\in[M]},\forall t \in [T]$ transmitted by $M$ codewords in the training sequence at each round to construct the channel decoder $\mathbf{G}_t\in\mathcal{G}$ with minimal expected error probability for each round. In this paper, we assume that $\mathcal{G}=\{\mathbf{G} \in \mathbb{R}^{d\times d}: \Vert \mathbf{G} \Vert_F \leq D \}$.

Given that the transceiver is equipped with channel output samples transmitted by $M$ codewords in the training sequence  related to the channel gain $\mathbf{H}_t$ and the channel noise $W_t$ at each round, we can calculate the empirical symbol error rate of the channel decoder $\mathbf{G}_t$ in the $t$-th round  using the training sequence, which includes $M$ distinct codewords as:
\begin{equation}\label{eq:fadingloss}
    \ell_t(\mathbf{G}_t):=\frac{1}{M}\sum_{j=1}^{M} \mathbbm{1}\Big \{\Vert \mathbf{x}^{j_t}-\mathbf{G}_t\mathbf{y}_t^j\Vert_{2}^2< \Vert \mathbf{x}^j-\mathbf{G}_t\mathbf{y}_t^j\Vert_{2}^2 \Big\}.
\end{equation}

Based on the above analysis, we notice that if a learned decoder $\mathbf{G}_t$ shows lower $\mathbb{P}(\mathbf{G}_t)$ for the training sequence, it can improve the transmission performance of data symbols. Hence, the transceiver can utilize the symbol error rate defined in Eq.~\eqref{eq:fadingloss} to carry out the online optimization protocol for learning the channel decoder $\mathbf{G}_t$ to enhance the transmission performance of data symbols. We design the online convex optimization algorithms to learn $\mathbf{G}_t$ in Section~\ref{sec:symobl}.

\subsubsection{Time-correlated Additive Noise Channels}
We then focus on the second specific channel model considered in this paper, and introduce the corresponding online optimization problem. For this channel, the channel transformation is assumed to be the identity mapping, i.e., $f_t=\mathbf{I}$,  so Eq.~\eqref{eq:channel}
becomes
\begin{equation}
    Y_t=X_t+Z_t,
\end{equation}
where $Z_t\in\mathbb{R}^{d}$ is a time-correlated channel noise with the distribution $F_{Z_t}$, which is statistically independent of the input. Similarly, we do not make any assumptions on the distribution of $Z_t$,  and $\{F_{Z_t}\}_{t \in [T]}$ is completely unknown to the transceiver. We suppose that the channel noise is time-correlated, i.e.,  $Z_t$ is dependent on $\{Z_{\tau}\}_{\tau \in [t-1]}$ from the previous round. We can also consider a first-order Markov noisy channel: $Z_{t+1}=\sqrt{1-\mu_t}Z_{t}+\sqrt{\mu_t}\boldsymbol{\epsilon}_t$, where $\boldsymbol{\epsilon}_t \in \mathbb{R}^{d}$ is a random vector and $\mu_t \in [0,1]$. This channel is pertinent to practical communication scenarios, where the presence of thermal noise in the receiver leads to gradual variations.

For this channel, we fix the channel decoder $\mathbf{G}_t=I$ for each round since the channel transformation $f_t$ is an identity function in this scenario.  Similar to the time-correlated fading channel, we also choose the nearest neighbor decoding rule based on the Euclidean distance, i.e., given a channel output $\mathbf{y}_t$, the index of the decoded codeword is chosen as:
\begin{equation}
    \begin{aligned}
        j (\mathbf{y}_t)\in \mathop{\arg \min}\limits_{j' \in[M]}\Vert \mathbf{x}_t^{j'}-\mathbf{y}_t \Vert_2,
    \end{aligned}
\end{equation}
where $\mathbf{y}_t=X_t+Z_t$ and the data symbol $X_t$ is uniformly sampled from the codebook $C_t=\{\mathbf{x}_t^j\}_{j\in [M]}$.
Notice that we do not assume that the time-correlated noise is Gaussian and this decoding rule is selected for its simplicity~\cite{weinberger2021generalization,DBLP:journals/tit/TsvieliW23}.

Analogously, as shown in Fig~\ref{fig:system} (b),  the transmitter sends $M$ distinct codewords in the deterministic training sequence and random data symbols to the receiver in each round.  We also define the expected error probability over the $t$-th round channel noise $Z_t\sim F_{Z_t}$ for $M$ distinct codewords in the training sequence as
\begin{equation}\label{eq:codebookprob}
\begin{aligned}
      & \mathbb{P}_{t}(C_t):=\frac{1}{M}\sum_{j=1}^{M} P\Big\{\Vert \mathbf{x}_t^{j_t}-\mathbf{y}_t^j\Vert_2^2<\Vert \mathbf{x}_t^j-\mathbf{y}_t^j\Vert_2^2\Big\},\\
\end{aligned}
\end{equation}
where  $j_t:=\arg\min_{j'\in [M]\setminus j}\Vert \mathbf{x}_t^{j'}-\mathbf{y}_t^{j}\Vert_{2}^2$ and $\mathbf{y}_t^j=\mathbf{x}_t^j+Z_t, \forall j \in [M]$. $C_t=\{\mathbf{x}_t^j\}_{j\in [M]}$ denotes the codebook chosen in the $t$-th round.

Similarly, we  presume that  the training sequence and data symbols share the same noise statistics during a communication round in this scenario. The codebook learned by the training sequence can thus contribute to improving the transmission reliability of data symbols.  Since data symbols are uniformly sampled from the learned codebook $C_t$, we  know that $\mathbb{E}_{X_t}[Pr\{\hat{X}_t\neq X_t|X_t;C_t\}]=\mathbb{P}_t(C_t)$, where $\hat{X}_t$ denotes the prediction of the channel output transmitted  by data symbol $X_t$. The online optimization goal for this scenario is to utilize the channel output samples  $\{\mathbf{y}_t^j\}_{j\in[M]}, \forall t \in [T]$ transmitted by $M$ codewords in the training sequence for selecting the codebook $C_t\in \mathcal{C}\subset(\mathbb{R}^d)^M$ with minimal expected error probability at each round. 

Based on channel output samples transmitted by $M$ codewords in the training sequence related to $Z_t$, the symbol error rate of codebook $C_t$ can be defined as
\begin{equation}\label{eq:correloss}
    \ell_t(C_t):=\frac{1}{M}\sum_{j=1}^{M} \mathbbm{1} \Big\{\Vert \mathbf{x}_t^{j_t}-\mathbf{y}_t^j\Vert_{2}^2< \Vert \mathbf{x}^j_t-\mathbf{y}_t^j\Vert_{2}^2\Big\}.
\end{equation}

Similarly, the transceiver can make use of this symbol error rate as the loss function to perform the online optimization procedure for learning the codebook $C_t$ to enhance the transmission performance of data symbols. We devise the multi-armed bandit algorithm to learn $C_t$ in Section~\ref{sec:codebook}.



\subsection{Online Optimization Procedure}
As mentioned above, this paper models the problem of 
learning the channel decoder $\mathbf{G}_t$ or codebook $C_t=\{\mathbf{x}_t^{j}\}_{j \in [M]}$ (a constellation)
as an online optimization problem.
Assume that there is a transmitter-receiver pair, and 
the transmitter sends  a package containing both the deterministic training sequence including $M$ codewords and random data symbols to the receiver over the time-correlated channel. Let $\mathcal{D}$ be a  feasible set
and $\ell:\mathcal{D}\rightarrow\mathbb{R}$ be a loss function.
In general,
at round $t\in [T]$, the transceiver carries out the following step:

\begin{enumerate}
    \item The receiver makes a decision $D_t\in \mathcal{D}$, 
            which can be either a channel decoder $\mathbf{G}_t$ or a codebook $C_t$. Following~\cite{weinberger2021generalization}, we also assume that the receiver can transmit the learned codebook $C_t$ to the  transmitter via the reliable feedback link given that communication resources of the receiver are sufficient.
    \item The transmitter then sends codewords to the receiver over the  time-correlated channel,  and then the receiver calculates the  symbol error rate based on  received channel outputs transmitted by $M$ codewords in the training sequence as the  loss function $\ell_t(D_t)$.
    \item The receiver leverages  $\ell_t(D_t)$ to run the online optimization algorithm for making the next decision.
\end{enumerate}

The objective of the considered online optimization problem is to construct a sequence of decisions $\{D_t\}_{t\in[T]}$, 
which minimizes the regret over $T$ rounds, defined as
\begin{equation}\label{eq:vanillareg}
\mbox{Reg}_T:=\sum_{t=1}^T\Big(\ell_t(D_t)-\ell_t(D^*)\Big),
\end{equation}
where $D^*:=\arg\min_{D \in \mathcal{D} }\sum_{t=1}^T \ell_t(D)$.

Subsequently, we will show the relationship between  minimizing this regret and minimizing the average expected error probability denoted by $\frac{1}{T}\sum_{t=1}^T \mathbb{P}_{t}(D_t)$. 
Leveraging this insight, 
we design various algorithms within a general optimistic online mirror descent (OMD) framework~\cite{DBLP:journals/jmlr/ChiangYLMLJZ12,DBLP:conf/colt/RakhlinS13} to 
learn  channel decoders or codebooks for communication over time-correlated channels, backed by solid theoretical guarantees. The benefit of this framework is to utilize the distribution dependence within predictable environments across different rounds for improving  online optimization procedures, which is suitable for time-correlated channels considered in this paper. 

Notice that we do not delve into the computational complexity 
or practical applications  of proposed algorithms in this paper. 
Our   focus is to explore theoretical performance limits of communication systems learned by devised algorithms.



\section{Learning Channel Decoder via Online Convex Optimization }\label{sec:symobl}
In this section, we consider the time-correlated fading channel and  fix the codebook $C_t=C=\{\mathbf{x}^j\}_{j\in[M]},\forall t \in [T]$  with the constant modulus constraint $\Vert \mathbf{x}^j\Vert_2=\gamma_X, \forall j\in[M]$~\cite{he2022qcqp} for each round. Hence, we only focus on designing algorithms to learn channel decoders $\{\mathbf{G}_t\}_{t\in[T]}$ to minimize the expected error probability defined in Eq.~\eqref{eq:fadingprob}. To conserve space, the proofs for all theorems below are provided in the appendix.

\subsection{Preliminaries}
In this subsection, we first introduce some vital physical quantities and typical assumptions related to the considered channel.   Next, we introduce a hinge-type surrogate loss and leverage it to solve the online decoder learning problem.

At first, we define the variance of the time-correlated channel gain $\{\mathbf{H}_t\}_{t\in[T]}$  as
\begin{equation}(\sigma_{\mathbf{H}_t})^2:=\mathbb{E}_{\mathbf{H}_t \sim F_{\mathbf{H}_t}}\Big[\Vert \mathbf{H}_t-\mathbf{U}_t\Vert_F^2| \mathcal{F}_{t-1}\Big], 
\end{equation}where $\mathbf{U}_t:=\mathbb{E}_{\mathbf{H}_t \sim F_{\mathbf{H}_t}} [\mathbf{H}_t| \mathcal{F}_{t-1}]$ denotes the mean matrix of $\mathbf{H}_t$  and $\mathcal{F}_{t-1}$ denotes the $\sigma$-algebra generated by $(\mathbf{H}_1,\mathbf{H}_2,...,\mathbf{H}_{t-1})$. 

Then we define the cumulative variance of $\{\mathbf{H}_t\}_{t\in[T]}$ as 
\begin{equation}\label{eq:variance}
    (\sigma^\mathbf{H}_{1:T})^2:=\mathbb{E}\Big[\sum_{t=1}^T(\sigma_{\mathbf{H}_t})^2\Big].
\end{equation}


Furthermore, we  define the cumulative  variation of time-correlated channel gain $\{\mathbf{H}_t\}_{t\in[T]}$ as
\begin{equation}
\label{eq:TCOM2024:cumulative_variation_cg}
(\Sigma^\mathbf{H}_{1:T})^2:=
\mathbb{E}\Big[\sum_{t=1}^T\Vert \mathbf{U}_t-\mathbf{U}_{t-1} \Vert^2_F \Big],
\end{equation}
where $\mathbf{U}_0=\mathbf{0}$. This quantity reflects the correlation of the channel gain between the previous and current rounds. We observe that $(\Sigma^\mathbf{H}_{1:T})^2$ decreases when $\{\mathbf{H}_t\}_{t\in[T]}$ exhibit high interdependence across different rounds.

Additionally, the following standard assumption widely used in the theoretical analysis of learning-based communication systems~\cite{weinberger2021generalization,DBLP:journals/tit/TsvieliW23} is required.

\begin{assumption}[Channel gain and channel noise with bounded expected norm]\label{assump:env}
The expected norm of channel gain is bounded by $\gamma_H$ and the expected norm of additive Gaussian noise is bounded by $\gamma_W$  at each round, i.e., for any $t\in[T]$, we have $\mathbb{E}\Vert \mathbf{H}_t\Vert_F\leq \gamma_H$ and $\mathbb{E}\Vert W_t\Vert_2=\sigma_W\leq \gamma_W$.
\end{assumption}


Based on this assumption, we  find that  $\{\mathbf{y}_t^j\}_{j \in [M],t\in [T]}$  are bounded by $\sqrt{2(\gamma_X\gamma_{H})^2+2(\gamma_W)^2}$ and we denote it by $L$ in the following.  

We then provide a convex surrogate loss for designing  algorithms to learn the channel decoder $\mathbf{G}_t$ via upper-bounding the expected error probability  in Eq.~\eqref{eq:fadingprob}. Notice that we know the inequality $\mathbbm{1}\{t<0\} \leq [r-t]_{+}$ holds, where $r\geq 1$. Hence, the expected error probability can be bounded as follows 
\begin{equation}\label{eq:union}
    \begin{aligned}
     \mathbb{P}_t(\mathbf{G}_t)&=\sum_{j=1}^M\frac{P\big\{\exists j'\neq j: \Vert \mathbf{x}^{j'}-\mathbf{G}_t\mathbf{y}_t^{j}\Vert^2_{2}\leq \Vert \mathbf{x}^{j}-\mathbf{G}_t\mathbf{y}_t^{j}\Vert^2_{2} \big\}}{M} \\
       & \stackrel{\text{(a)}}{\leq} \sum_{j=1}^M\sum^M_{j'\neq j} \frac{P\big\{\Vert \mathbf{x}^{j'}-\mathbf{G}_t\mathbf{y}_t^{j}\Vert^2_{2}<\Vert \mathbf{x}^{j}-\mathbf{G}_t\mathbf{y}_t^{j}\Vert^2_{2} \big\}}{M}\\
         &\stackrel{\text{(b)}}{\leq}\sum_{j=1}^M\sum^M_{j'\neq j} \frac{\mathbb{E}\big[r-\Vert \mathbf{x}^{j'}-\mathbf{G}_t\mathbf{y}_t^{j}\Vert^2_{2}+\Vert \mathbf{x}^{j}-\mathbf{G}_t\mathbf{y}_t^{j}\Vert^2_{2}\big]_{+}}{M},
    \end{aligned}
\end{equation}
where $r\geq 1$.  (a) holds based on the Boole's inequality. (b) follows from the fact that $\mathbb{E}[\mathbbm{1}\{\Vert \mathbf{x}^{j'}-\mathbf{G}_t\mathbf{y}_t^{j}\Vert^2_{2}-\Vert \mathbf{x}^{j}-\mathbf{G}_t\mathbf{y}_t^{j}\Vert^2_{2}\leq 0\}] \leq \mathbb{E}\big[r-\Vert \mathbf{x}^{j'}-\mathbf{G}_t\mathbf{y}_t^{j}\Vert^2_{2}+\Vert \mathbf{x}^{j}-\mathbf{G}_t\mathbf{y}_t^{j}\Vert^2_{2}\big]_{+}$ when $r$ satisfies $r\geq 1$, which can be found in~\cite{cesa2006prediction}.

We observe that $r-\Vert \mathbf{x}^{j'}-\mathbf{G}_t\mathbf{y}_t^{j}\Vert^2_{2}+\Vert \mathbf{x}^{j}-\mathbf{G}_t\mathbf{y}_t^{j}\Vert^2_{2}=r+2\langle\mathbf{x}^{j'}-\mathbf{x}^j,\mathbf{G}_t\mathbf{y}_t^{j}\rangle$ is an affine function w.r.t. $\mathbf{G}_t$, thus $[r-\Vert \mathbf{x}^{j'}-\mathbf{G}_t\mathbf{y}_t^{j}\Vert^2_{2}+\Vert \mathbf{x}^{j}-\mathbf{G}_t\mathbf{y}_t^{j}\Vert^2_{2}]_{+}$ is a convex function w.r.t. $\mathbf{G}_t$. Based on the above observation, we utilize the  following hinge-type loss as the convex surrogate loss function with respect to $\mathbf{G}_t$ with the parameter $r_t\geq 1$:
 \begin{equation}\label{eq:ocoproblem}
\begin{aligned}\Tilde{\ell}_t(\mathbf{G}_t):=\sum_{j=1}^M\sum^M_{j'\neq j}\frac{\big[r_t-\Vert \mathbf{x}^{j'}-\mathbf{G}_t\mathbf{y}_t^{j}\Vert^2_{2}+\Vert \mathbf{x}^{j}-\mathbf{G}_t\mathbf{y}_t^{j}\Vert^2_{2}\big]_{+}}{M}.\\
\end{aligned}
\end{equation}


The subgradient of this surrogate loss function  $\Tilde{\ell}_t(\mathbf{G}_t)$ is
\begin{equation}\label{eq:subgradient}
\begin{aligned}
\nabla_{\mathbf{G}_t}\Tilde{\ell}_t(\mathbf{G}_t)
       &=\frac{2}{M}\sum_{j=1}^M \sum^M_{j'\neq j} \mathbbm{1}^{j,j'}_t(r_t) (\mathbf{x}^{j'}-\mathbf{x}^j)(\mathbf{y}_t^j)^T,
\end{aligned}
\end{equation}
where $\mathbbm{1}^{j,j'}_t(r_t):=\mathbbm{1}\{\Vert \mathbf{x}^{j'}-\mathbf{G}_t\mathbf{y}_t^{j}\Vert^2_{2}-\Vert \mathbf{x}^{j}-\mathbf{G}_t\mathbf{y}_t^{j}\Vert^2_{2}\leq r_t\}$.

Based on the proposed surrogate loss, we can redefine the regret defined in Eq.~\eqref{eq:vanillareg} for this scenario as follows,
\begin{equation}\label{eq:reg}
\mbox{Reg}'_T:=\sum_{t=1}^T\Big(\Tilde{\ell}_t(\mathbf{G}_t)-\Tilde{\ell}_t(\mathbf{G}^*)\Big),
\end{equation}
where $\mathbf{G}^*:=\arg\min_{\mathbf{G}\in \mathcal{G}}\sum_{t=1}^T\Tilde{\ell}_t(\mathbf{G})$. 

\subsection{Optimistic OMD with Euclidean Regularizer}
Using the proposed hinge-type surrogate loss, we can regard the online channel decoder learning problem as an online convex optimization problem minimizing the regret defined in Eq.~\eqref{eq:reg}. In this section, we devise an online optimization algorithm based on the  optimistic OMD framework to learn the channel decoder $\mathbf{G}_t$ on the fly. 


Specifically,  during the online optimization process, the transceiver stores two sequences $\{\mathbf{G}_t\}_{t=1}^T$ and $\{\mathbf{G}'_t\}_{t=1}^T$. At each round $t\in [T]$,  the transceiver initially uses a hint matrix $\mathbf{M}_{t}\in \mathcal{G}$, which incorporates specific prior knowledge of the unknown channel gain $\mathbf{H}_{t}$, to construct 
the channel decoder $\mathbf{G}_t$. Then,  the transceiver utilizes the learned $\mathbf{G}_t$ to finish one round of communication and calculates the corresponding surrogate loss $\Tilde{\ell}_t(\mathbf{G}_t)$.

Then we introduce the procedure of optimistic OMD~\cite{DBLP:conf/colt/RakhlinS13} below, which is
defined as the following two step updates 
\begin{equation}\label{eq:OMD}
    \begin{aligned}
\mathbf{G}_{t}&=\underset{\mathbf{G}\in\mathcal{G}}{\arg\min}  \big\{\langle \mathbf{M}_{t},\mathbf{G}\rangle+\mathcal{B}_{\psi_{t}}(\mathbf{G},\mathbf{G}'_{t})\big\},\\
  \mathbf{G}'_{t+1}&=\underset{\mathbf{G}\in\mathcal{G}}{\arg\min}  \big\{\langle\nabla\Tilde{\ell}_t(\mathbf{G}_t),\mathbf{G}\rangle+\mathcal{B}_{\psi_{t}}(\mathbf{G},\mathbf{G}'_{t})\big\},
    \end{aligned}
\end{equation}
where $\mathcal{B}_{\psi}(X,Y)=\psi(X)-\psi(Y)-\langle\nabla\psi(Y),X-Y\rangle$ denotes the Bregman divergence induced by a differentiable convex function $\psi$, which is called the regularizer. We allow the regularizer $\psi_t$ to be time-varying in this paper. In essence, by performing twice mirror descent, our algorithm exploits channel correlation to enhance online optimization processes.



Given that the channel gain $\{\mathbf{H}_t\}_{t\in [T]}$ 
are mutually dependent across different rounds in this scenario, 
we set the hint matrix $\mathbf{M}_{t}$ as $\nabla\Tilde{\ell}_{t-1}(\mathbf{G}_{t-1})$,
i.e., the last-round gradient,
to enhance the online optimization process.  
In addition, we can set $\mathbf{G}_1=\mathbf{G}'_1$ to be an arbitrary matrix in $\mathcal{G}$.

In this section, we set the regularizer as the Euclidean norm~\cite{DBLP:conf/icml/ChenT0023}, i.e., $\psi_t(\mathbf{G})=\frac{1}{2\eta_t}\Vert \mathbf{G}\Vert^2_F$ with the learning rate $  \eta_t=\frac{D}{\sqrt{1+\sum_{\tau=1}^{t-1}\Vert \nabla\Tilde{\ell}_{\tau}( \mathbf{G}_{\tau})
    -\mathbf{M}_{\tau}\Vert_F^2}}$.

 To sum up, the update rules in Eq.~\eqref{eq:OMD} become
\begin{equation}
    \begin{aligned}
    \mathbf{G}_{t}&=\Pi_{\mathcal{G}}[  \mathbf{G}'_{t}-\eta_{t}\nabla\Tilde{\ell}_{t-1}( \mathbf{G}_{t-1})],\\
             \mathbf{G}'_{t+1}&=\Pi_{\mathcal{G}}[  \mathbf{G}'_{t}-\eta_t\nabla\Tilde{\ell}_t(\mathbf{G}_t)],\\
    \end{aligned}
\end{equation}
where $\Pi_{\mathcal{G}}$ denotes the Euclidean projection onto the feasible
domain $\mathcal{G}$. In fact, the proposed approach performs gradient descent twice at each round.  Besides, the step size $\{\eta_t\}_{t\in[T]}$ is chosen adaptively, 
similar to self-confident tuning \cite{Auer2002Adaptive}.

\begin{algorithm}[h]
 \KwIn{ The number of communication round $T$, step size $\{\eta_t\}_{t\in[T]}$, the parameters $\{r_t\}_{t\in[T]}$ in surrogate loss.}
   \textbf{Initialize:} $\mathbf{G}_1\in \mathcal{G}$, and  $\mathbf{G}_1=\mathbf{G}'_1$ \;
  \For{round  $t \in [T]$ }{
 Update the  channel decoder: $     \mathbf{G}_{t}=\Pi_{\mathcal{G}}[ \mathbf{G}'_{t}-\eta_{t}\nabla\Tilde{\ell}_{t-1}(\mathbf{G}_{t-1})]$\;
   The transmitter sends codewords to the receiver\;
 The receiver calculates $\Tilde{\ell}_t(\mathbf{G}_t)$\;
 Update the auxiliary channel decoder:  $     \mathbf{G}'_{t+1}=\Pi_{\mathcal{G}}[ \mathbf{G}'_{t}-\eta_t\nabla\Tilde{\ell}_t(\mathbf{G}_t)]$\;
  }
  \caption{Optimistic OMD for Learning Channel decoder}
  \label{alg1}
\end{algorithm}

The protocol of the proposed algorithm is illustrated in Algorithm~\ref{alg1}. In practice, we can adaptively modify the filter coefficients of a linear discrete-time filter according to the $\mathbf{G}_t$ learned by the optimistic OMD algorithm. Since adaptive equalization is a widely used technique in modern communication systems,  our algorithm can be feasibly deployed in real-world systems. In the following, we establish the theoretical guarantee of the proposed algorithm via offering an upper bound on the expected error probability with learned channel decoders $\{\mathbf{G}_t\}_{t\in[T]}$.

\begin{theorem}\label{thm:oco} Under Assumption~\ref{assump:env}, if we select the parameters $r_t=2d^*DL+\frac{1}{\sqrt{T}}$ with the maximum codewords distance $d^*:=\max_{i\neq j}\Vert \mathbf{x}^i-\mathbf{x}^j\Vert_2\geq\frac{1}{2DL}$, we have
    \begin{equation}
        \begin{aligned}
           & \frac{1}{T}\sum_{t\in[T]}\mathbb{P}_t(\mathbf{G}_t)-\frac{1}{T}\sum_{t\in[T]}\mathbb{E}[\Tilde{\ell}_{t}(\mathbf{G}^*)]\\
&\leq\mathcal{O}\Big(\frac{1}{T}\sqrt{M\sum_{t\in [T]}\sum_{j\in [M]}\mathbb{E}\big\Vert
    \mathbf{y}_t^j-\mathbf{y}_{t-1}^j\big\Vert_2^2}\Big)\\
    &\leq \mathcal{O}
       \Big(  \frac{M}{T}\sqrt{\sum_{t\in[T]}\Big[\gamma_X^2\Big(\gamma_H^2-\mathbb{E}[\langle\mathbf{H}_{t},\mathbf{H}_{t-1}\rangle]\Big)+\sigma_W^2\Big]}\Big).
        \end{aligned}
    \end{equation}
    \begin{remark}
  Theorem~\ref{thm:oco}  provides a performance guarantee of the learned  channel decoder $\{\mathbf{G}_t\}_{t\in[T]}$: if $\exists \mathbf{G}^* \in \mathcal{G}$ such that the term $\sum_{t\in[T]}\mathbb{E}[\Tilde{\ell}_{t}(\mathbf{G}^*)]$ is sub-linear w.r.t $T$, i.e., $\sum_{t\in[T]}\Tilde{\ell}_t(G^*)=\mathcal{O}(T^{\alpha}),\alpha<1$,  the average expected error probability satisfies $\frac{1}{T}\sum_{t\in[T]}\mathbb{P}_t(\mathbf{G}_t) \rightarrow 0$ as $T \rightarrow \infty$.  This is because $\mathbb{E}\Vert \mathbf{y}_t^j-\mathbf{y}_{t-1}^j\Vert_2^2\leq 4L^2$, and we  have $ \frac{1}{T}  \sqrt{M\sum_{t\in[T]}\sum_{j\in[M]}\mathbb{E}\Vert \mathbf{y}_t^j-\mathbf{y}_{t-1}^j\Vert_2^2}\leq \frac{2ML\sqrt{T}}{T}=\mathcal{O}(\frac{\sqrt{T}}{T})$. Moreover, we emphasize that this upper bound is meaningful when $\mathbb{P}_t(\mathbf{G}_t) \leq \mathbb{E}\tilde{\ell}_t(\mathbf{G}^*), \forall t \in [T]$, as in this case  $\frac{1}{T} \sum_t \mathbb{P}_t(\mathbf{G}_t) \leq \frac{1}{T} \sum_t \mathbb{E} \tilde{\ell}_t(\mathbf{G}^*)$, which vanishes as $T \rightarrow \infty$.
  
We notice that a larger positive auto-correlation function $\mathbb{E}[\langle\mathbf{H}_{t},\mathbf{H}_{t-1}\rangle]$ of $\{\mathbf{H}\}_{t\in[T]}$ results in a tighter derived upper bound.  This indicates that the proposed method  utilizes the channel correlation (second-order channel statistics) to improve the performance of learned decoder.   For the practical implementation of our method, we do not need to tune $r_t$ based on the prior knowledge of $L$ and $D$. Instead, we  compute the sub-gradient  as  $\nabla \tilde{\ell}_t(\mathbf{G}_t) = \frac{2}{M}\sum_{j\in[M]}\sum_{j'\neq j}(\mathbf{x}^{j'} - \mathbf{x}^{j})(\mathbf{y}_t^j)^T$ used to update decoders,  corresponding to set $r_t=2d^*DL+\frac{1}{\sqrt{T}}$.
    \end{remark}
\end{theorem}

 To better understand the result of Theorem~\ref{thm:oco}, we utilize the physical quantities introduced before to provide the below corollary based on Theorem~\ref{thm:oco}.

\begin{corollary}\label{corollary:oco} Under the conditions of Theorem~\ref{thm:oco}, we have
        \begin{equation}
        \begin{aligned}
           & \frac{1}{T}\sum_{t\in[T]}\mathbb{P}_t(\mathbf{G}_t)-\frac{1}{T}\sum_{t\in[T]}\mathbb{E}[\Tilde{\ell}_{t}(\mathbf{G}^*)]\\
           &\leq\mathcal{O}\Big(\frac{M}{T}\sqrt{(\sigma^{\mathbf{H}}_{1:T})^2}+\frac{M}{T}\sqrt{(\Sigma^{\mathbf{H}}_{1:T})^2}+\frac{M\sqrt{T}}{T}\sqrt{\sigma_W^2}\Big),
        \end{aligned}
    \end{equation}
where 
    $(\sigma^{\mathbf{H}}_{1:T})^2$ follows 
    Eq. \eqref{eq:variance}
    and 
    $(\Sigma^{\mathbf{H}}_{1:T})^2$ follows 
    Eq. \eqref{eq:TCOM2024:cumulative_variation_cg}.
\end{corollary}

\begin{remark}
    Corollary~\ref{corollary:oco} implies that the variance $\sigma^2_W$ of the channel noise $W$, the cumulative variance $(\sigma^{\mathbf{H}}_{1:T})^2$ and the cumulative variation $(\Sigma^{\mathbf{H}}_{1:T})^2$ of channel gain $\{\mathbf{H}_t\}_{t\in[T]}$ can deteriorate the performance of  the learned channel decoder $\{\mathbf{G}_t\}_{t\in [T]}$. We observe that  smaller $(\Sigma^{\mathbf{H}}_{1:T})^2$ leads to a tighter upper bound of average expected error probability $\frac{1}{T}\sum_{t\in[T]}\mathbb{P}_t(\mathbf{G}_t)$, reflecting that the proposed method utilizes the distribution dependence of the time-correlated channel gain $\{\mathbf{H}_t\}_{t\in[T]}$ to enhance the online optimization process. Based on Assumption~\ref{assump:env}, we know that this upper bound becomes $ \mathcal{O}\left(\frac{\gamma_H M \sqrt{T}}{T} + \frac{M\sqrt{T}}{T}\sqrt{\sigma_W^2}\right)$, which vanishes as $T \to \infty$.
\end{remark}

Lastly, we briefly discuss practical applications of our method. For example, in MIMO systems with high-mobility users, our proposed  optimistic OMD method enables dynamic precoding matrix updates via pilot signals, significantly enhancing overall transmission performance.

\section{ Learning codebook via Multi-armed Bandit}\label{sec:codebook}
In this section, we focus on the time-correlated additive noise channel and set the channel decoder $\{\mathbf{G}_t\}_{t\in [T]}$ as the identity matrix $\mathbf{I}$. Hence, we only consider devising algorithms to learn the codebook $\{C_t\}_{t\in[T]}$  to minimize the expected error probability defined in Eq.~\eqref{eq:codebookprob}. Similarly, for brevity, the proofs for all theorems below are available in the appendix.

\subsection{Preliminaries}
Generally, identifying an optimal codebook becomes challenging when the noise distribution is unknown~\cite{weinberger2021generalization,liu2024pac}. Even more complex is the dynamic adjustment of the codebook to accommodate channel noise  with the   time-varying  distribution.

One potential approach to tackle this challenge involves selecting codebooks for practical transmission from a pre-defined super-codebook $\mathbf{C}=\{C_i\}_{i=1}^N$ comprising $|\mathbf{C}|=N$ codebooks~\cite{weinberger2021generalization,DBLP:journals/jsac/ForneyW89}. This super-codebook can be statically constructed (such as a grid or a lattice~\cite{zamir2014lattice}) in advance. To adapt to the changing statistical property of the channel noise, we dynamically choose one of these codebooks used for transmitting codewords in each round. Specifically, we select the codebook based on the  symbol error rate of codebooks over the channel noise $Z_t,\forall t\in[T]$.   This approach draws inspiration from the Gibbs-algorithm-based codebook expurgation proposed in~\cite{weinberger2021generalization}, and is amenable to practical deployment following the paradigm of Adaptive Modulation and Coding (AMC) widely deployed in communication systems~\cite{goldsmith2005wireless}. Based on it,  we consider that the  super-codebook $\mathbf{C}$ is known to both the transmitter and receiver. Hence, the receiver  is not required to transmit the entire learned codebook to the transmitter.  Instead, the receiver only needs to send the index of the codebook selected in this round to the transmitter, thereby reducing the cost of executing online codebook learning.

 In this paper, we model this online codebook learning problem as a multi-armed bandit problem~\cite{cesa2006prediction}. We consider an iterative process spanning $T$ rounds below. In each round, the transceiver selects a codebook $C_t $ from the super-codebook $\mathbf{C}$ and uses it for transmitting codewords over $Z_t$. Then the receiver calculates the corresponding  symbol error rate of this chosen codebook $C_t$ as the loss. We denote codewords in $C_t$ as $\{\mathbf{x}_t^j\}_{j\in[M]}$, and define the loss  of $C_t$ as
 \begin{equation}\label{eq:codebookloss}
     \ell_t(C_t):=\frac{1}{M}\sum_{j\in[M]}\mathbbm{1}\Big\{\Vert \mathbf{x}_t^{j_t} -\mathbf{y}_t^j\Vert_2^2\leq \Vert \mathbf{x}_t^{j}-\mathbf{y}_t^j\Vert_2^2\Big\},
 \end{equation}
 where $\mathbf{y}_t^j=\mathbf{x}^j_t+Z_t$ and $j_t=\arg\min_{j'\in[M]\setminus j }\Vert \mathbf{x}_t^{j'} -\mathbf{y}_t^j\Vert_2^2$.

 Formally, the transceiver selects a binary vector $\mathbf{a}_t$ called the index vector from the feasible set $\mathcal{X}:=\{\mathbf{e}_{1},\mathbf{e}_{2},...,\mathbf{e}_{N}\}$, where $\mathbf{e}_{i}$ denotes the $i$-th standard basis vector. In other words, in each round, the transceiver chooses the index $i_t \in [N]$ (corresponding to $\mathbf{a}_t=\mathbf{e}_{i_t}$) of the codebook $C_t$. The transceiver then uses $C_t$ for transmitting codewords, and suffers loss denoted by $\mathbf{a}_t^T\boldsymbol{\ell}_t=\ell_t(C_t)$, where $\boldsymbol{\ell}_t\in [0,1]^N$ is a vector including all the  symbol error rates of codebooks in $\mathbf{C}$ transmitted over $Z_t$. The regret now can be redefined as
 \begin{equation}
\mbox{Reg}_T:=\sum_{t\in[T]}\left[\mathbf{a}_t^T\boldsymbol{\ell}_t-(a^*)^T\boldsymbol{\ell}_t\right]
    =\sum_{t\in[T]}\left[\ell_t(C_t)-\ell_t(C^*)\right],
 \end{equation}
where $\mathbf{a}^*:=\min_{\mathbf{a}\in\mathcal{X}}\sum_{t\in[T]}\mathbf{a}^T\boldsymbol{\ell}_t$ and $C^*\in \mathbf{C}$ denotes the codebook corresponding to the index vector $\mathbf{a}^*$.

\subsection{Optimistic OMD with Log-Barrier Regularizer}

In this section, we also use the optimistic OMD framework to design an online optimization algorithm for solving the problem of selecting codebooks to communicate over time-correlated additive noise channels. Similarly, the proposed algorithm offers the advantage of utilizing the distribution dependence of such channels for boosting bandit learning processes,  suited for the channel considered in this section. 

The  OMD framework employed in  bandit operates on the set $\Omega=\mbox{conv}(\mathcal{X}):=\{\sum_{i\in[N]}\beta_i \mathbf{e}_i:\sum_{i\in[N]}\beta_i=1,\beta_i\geq 0,\forall i\in[N]\}$. The update rule of OMD for bandit is $\mathbf{w}_t=\arg\min_{\mathbf{w}\in \Omega}\{\langle \mathbf{w},\hat{\boldsymbol{\ell}}_{t-1}\rangle+\mathcal{B}_{\psi}(\mathbf{w},\mathbf{w}_{t-1})\}$ for the regularizer $\psi$ and an unbiased estimator $\hat{\boldsymbol{\ell}}_{t-1}$ of the true loss $\boldsymbol{\ell}_{t-1}$. The transceiver then selects the index vector $\mathbf{a}_t$ randomly such that $\mathbb{E}[\mathbf{a}_t]=\mathbf{w}_t$, which corresponds to the codebook $C_t$. In essence, $\mathbf{a}_t$ is sampled from the probability distribution $\mathbf{w}_t$. Then we construct  the next $\hat{\boldsymbol{\ell}}_t$ based on the feedback. 

In this section,  the optimistic OMD framework also involves maintaining a sequence of auxiliary action $\mathbf{w}'_t$ updated by $\hat{\boldsymbol{\ell}}_t$. As mentioned above, Optimistic OMD makes a decision $\mathbf{a}_t\sim \mathbf{w}_t$ randomly, and $\mathbf{w}_t$ is now updated by minimizing  $\mathbf{m}_t \in [0,1]^{N}$, an optimistic hint of the true loss $\boldsymbol{\ell}_t$.  Hence,  the update rules of optimistic OMD for this scenario become 
\begin{equation}
    \begin{aligned}
   \mathbf{w}_{t}    &= \mathop{\arg \min}\limits_{\mathbf{w}\in\Omega}\big\{\langle \mathbf{w}, \mathbf{m}_{t}  \rangle+\mathcal{B}_{\psi_{t} }(\mathbf{w},\mathbf{w}'_{t})\big\},\\
      \mathbf{w}'_{t+1}    &= \mathop{\arg \min}\limits_{\mathbf{w}\in\Omega}\big\{\langle \mathbf{w}, \hat{\boldsymbol{\ell}}_t  \rangle+\mathcal{B}_{\psi_{t} }(\mathbf{w},\mathbf{w}'_{t})\big\}.\\
    \end{aligned}
\end{equation}



Following~\cite{DBLP:conf/colt/WeiL18}, we set the regularizer as the Log-Barrier $\psi_t(\mathbf{w})=\sum_{i \in [N]}\frac{1}{\eta_t}\ln\frac{1}{\mathbf{w}_i}$ with 
learning rate $\eta_t$ for deriving our theoretical results.
Recall that we consider the time-correlated additive noise channel in this section, i.e.,  $Z_t$ depends on $\{Z_{\tau}\}_{\tau\in[t-1]}$ from previous rounds. Therefore, for utilizing such dependence to enhance the bandit learning process, we set the  $i$-th component $\mathbf{m}_{t,i}$ of $\mathbf{m}_t$ to be the most recent observed loss of codebook $i \in [N]$. Specifically, $\mathbf{m}_{t,i}$ is set as $\mathbf{m}_{t,i}=\boldsymbol{\ell}_{\alpha_i(t),i}$, where $\alpha_i(t)$ is defined to be the most recent time when codebook $i$ is chosen prior to round $t$, that is $\alpha_i(t):=\max\{\tau<t:i_{\tau}=i\}$ (or $0$ if the set is empty).

\begin{algorithm}[h]
 \KwIn{ The number of communication round $T$, step size $\{\eta_{t}\}_{t\in[T]}$.}
   \textbf{Initialize:} $\mathbf{w}'_1=\arg\min_{\mathbf{w}\in \Omega}\psi_1(\mathbf{w})$ \;
  \For{round  $t \in [T]$ }{
    Update the action:  $    \mathbf{w}_{t}    = \mathop{\arg \min}\limits_{w\in\Omega}\{\langle \mathbf{w}, \mathbf{m}_{t}  \rangle+\mathcal{B}_{\psi_{t} }(\mathbf{w},\mathbf{w}'_{t})\}$\;
  The transmitter sends codewords to the receiver based on the codebook $C_t$ corresponding to the index vector $\mathbf{a}_t\sim \mathbf{w}_t $\;
  The receiver calculates $\mathbf{a}_t^T\boldsymbol{\ell}_t=\ell_t(C_t)$ and constructs the unbiased estimator $\hat{\boldsymbol{\ell}}_t$ of $\boldsymbol{\ell}_t$\;
 Update the auxiliary action: $    \mathbf{w}'_{t+1}    = \mathop{\arg \min}\limits_{\mathbf{w}\in\Omega}\{\langle \mathbf{w},\hat{\boldsymbol{\ell}}_t\rangle+\mathcal{B}_{\psi_t}(\mathbf{w},\mathbf{w}_t')\}$\;
  }
  \caption{Optimistic OMD for Learning codebook}
  \label{alg3}
\end{algorithm}

The protocol of the presented  algorithm  for solving the online codebook learning problem under the time-correlated additive noise channel is illustrated in Algorithm~\ref{alg3}.

Analogously, we provide the theoretical guarantee for the proposed method via deriving an upper bound on the averaged expected
error probability of learned codebooks $\{C_t\}_{t\in [T]}$.

\begin{theorem}\label{thm:bandit} 
Let $\hat{\boldsymbol{\ell}}_t$ be an estimator of $\boldsymbol{\ell}_t$, satisfying 
$$
\forall i\in[N],\quad\hat{\boldsymbol{\ell}}_{t,i}
    =\frac{\boldsymbol{\ell}_{t,i}-\mathbf{m}_{t,i}}{\mathbf{w}_{t,i}}\cdot\mathbbm{1}\{i_t=i\}
    +\mathbf{m}_{t,i},
$$
and  set the learning rate $\eta_t\leq \frac{1}{162}$ using the doubling trick~\cite{DBLP:conf/colt/WeiL18,cesa2006prediction}, we have
    \begin{equation}
        \begin{aligned}
        &  \frac{1}{T}\sum_{t\in [T]}\mathbb{P}_t(C_t)-\frac{1}{T}\sum_{t\in [T]}\mathbb{E}[\ell_t(C^*)]\\
        &\leq \Tilde{\mathcal{O}}\Big(\frac{1}{T}\sqrt{\sum_{t\in[T]}\sum_{i\in[N]}\frac{1}{M}\sum_{j\in[M]}\big\vert\mathbb{P}^j_{t}(i)-\mathbb{P}^j_{t-1}(i) \big\vert}\\
        &\quad+ \frac{1}{T}\sqrt{\sum_{t\in[T]}\sum_{i\in[N]}\frac{1}{M}\sum_{j\in[M]}\sigma[\mathbbm{1}_t^{j}(i)]}\Big)\\
        &\leq \tilde{\mathcal{O}}(\sqrt{\frac{N}{T}}),
        \end{aligned}
    \end{equation}
    where  $\mathbbm{1}_t^{j}(i):=\mathbbm{1}\{\Vert \mathbf{x}_t^{j_t} -\mathbf{y}_t^j\Vert_2\leq \Vert \mathbf{x}_t^{j} -\mathbf{y}_t^j\Vert_2 | \mathbf{a}_t=\mathbf{e}_{i}\}$ denotes the indicator of misclassifying the $j$-th codeword when the $i$-th codebook is selected in round $t$. $\mathbb{P}^j_{t}(i):=\mathbb{E}[\mathbbm{1}_t^{j}(i)]$ denotes the expectation of   $\mathbbm{1}_t^{j}(i)$, and $\sigma[\mathbbm{1}_t^{j}(i)]:=\sqrt{\mathbb{E}[\mathbbm{1}_t^{j}(i)-\mathbb{P}^j_{t}(i)]^2}$ denotes the standard deviation of $\mathbbm{1}_t^{j}(i)$.  
    \begin{remark}
        Theorem~\ref{thm:bandit} presents  theoretical performance guarantees of $\{C_t\}_{t\in[T]} $: if $\exists 
    C^*\in\mathbf{C}$ such that $\sum_{t\in[T]}\ell_t(C^*)$ is sub-linear w.r.t $T$, i.e., $\sum_{t\in[T]}\ell_t(C^*)=\mathcal{O}(T^{\alpha}),\alpha<1$, then we have $\frac{1}{T}\sum_{t\in[T]} \mathbb{P}_t(C_t)\rightarrow 0$, as $T\rightarrow \infty$. The reason is provided as follows. Since $1_t^{j}(i)$ is a Bernoulli random variable, we have $\sigma[1_t^{j}(i)]\leq 1$ and $\vert\mathbb{P}^j_{t}(i)-\mathbb{P}^j_{t-1}(i) \vert\leq2$. Hence, we know $ \frac{1}{T} \sqrt{\sum_{t\in [T]}\sum_{i\in[N]}\sum_{j\in[M]}\frac{\big\vert\mathbb{P}^j_{t}(i)-\mathbb{P}^j_{t-1}(i) \big\vert}{M}}\leq \sqrt{\frac{2N}{T}}$ and $ \frac{1}{T} \sum_{t\in[T]}\sqrt{\sum_{i\in[N]}\sum_{j\in[M]}\frac{\sigma [1_t^{j}(i)]}{M}}\leq \sqrt{\frac{N}{T}}$.
    
    In addition, based on Pinsker's inequality~\cite{cesa2006prediction}, we have $\sum_{t\in[T]}\sum_{i\in[N]}\frac{1}{M}\sum_{j\in[M]}\big\vert\mathbb{P}^j_{t}(i)-\mathbb{P}^j_{t-1}(i) \big\vert\leq N\sum_{t\in[T]}\sqrt{D_{kl}(\mathbb{P}_{t}\Vert \mathbb{P}_{t-1})}  $, where $D_{kl}(\mathbb{P}_{t}\Vert \mathbb{P}_{t-1}):=\arg\max_{i\in[N],j\in[M]}D_{kl}(\mathbb{P}^j_{t}(i)\Vert \mathbb{P}^j_{t-1}(i))$ and $D_{kl}(P\Vert Q)=\mathbb{E}_P\log\frac{dP}{dQ}$ denotes the relative entropy between probability measures $P$ and $Q$. $D_{kl}(\mathbb{P}_{t}\Vert \mathbb{P}_{t-1})$ essentially  captures the discrepancy  between the channel distributions of successive rounds. We notice that a smaller discrepancy $D_{kl}(\mathbb{P}_{t}\Vert \mathbb{P}_{t-1})$ results in a tighter regret bound.  It indicates that we can learn the optimal codebook $C_t$ if the statistical properties of $Z_t$  remain relatively stable, implying that the proposed method leverages the distribution dependence of  $\{Z_t\}_{t\in [T]}$ to improve the bandit learning process.
    \end{remark}
\end{theorem}

\subsection{Case study: $2$-ary codebook for Gaussian channels}
To better understand the result from Theorem~\ref{thm:bandit}, we consider
the below example about utilizing $2$-ary codebook to transmit codewords over Gaussian channels. Specifically, we assume that the time-correlated channel noise $Z_t$ is a zero-mean Gaussian noise with the variance of $\sigma^2_{Z_t}$. Additionally, we consider the $2$-ary codebook below, i.e., we set the number $M$ of codewords satisfies $M=2$ and thus $C_i=\{ \mathbf{x}^1_i, \mathbf{x}^2_i\}, \forall i\in[N]$. Correspondingly, we denote the distance between the two codewords in the $i$-th codebook as $d_i:=\Vert \mathbf{x}^1_i-\mathbf{x}^2_i\Vert_2$.

Then we can directly calculate the expected error probability  of  the $i$-th codebook over $Z_t$  as 
\begin{equation}
    \mathbb{P}_{Z_t}(C_i)= Q\big(\frac{d_{i}}{2\sigma_{Z_t}}\big):=\int_{ \frac{d_{i}}{2\sigma_{Z_t}}}^{\infty}\frac{1}{\sqrt{2\pi}} \exp(-\frac{\tau^2}{2})d\tau.
\end{equation}

Based on it, we can derive the average expected error probability of the $2$-ary codebook learned by our method for this time-correlated Gaussian channel as follows.

 \begin{corollary}\label{thm:banditGaussian} Under the conditions of Theorem~\ref{thm:bandit}, if for any $t\in[T]$, $Z_t$ is a zero-mean Gaussian noise with the variance of $\sigma^2_{Z_t}$, and the number of codewords in any codebooks in $\mathbf{C}$ is set as two, i.e., $M=2$, we have
     \begin{equation}
         \begin{aligned}
            &  \frac{1}{T}\sum_{t\in [T]}\mathbb{P}_t(C_t)-\frac{1}{T}\sum_{t\in [T]}\mathbb{E}[\ell_t(C^*)]\\
           &\leq \Tilde{\mathcal{O}}\Big(\frac{1}{T} \sqrt{ \sum_{i\in[N]}\sum_{t\in[T]}\sqrt{Q\big(\frac{d_{i}}{2\sigma_{Z_t}}\big)\Big(1-Q\big(\frac{d_{i}}{2\sigma_{Z_t}}\big)\Big)}}\\
           &\quad + \frac{1}{T}\sqrt{ \sum_{i\in[N]}\sum_{t\in[T]}\big|\sigma_{Z_t}-\sigma_{Z_{t-1}}\big|}\Big).\\
         \end{aligned}
     \end{equation}

     \begin{remark}
         Corollary~\ref{thm:banditGaussian} indicates that smaller  difference $|\sigma_{Z_t}-\sigma_{Z_{t-1}}|$ of the standard deviation $\{\sigma_{Z_t}\}_{t\in[T]}$ between successive rounds makes the performance of the learned codebook $\{C_t\}_{t\in [T]}$ better, implying that the proposed method fully leverages the distribution dependence of time-correlated Gaussian noise $\{Z_t\}_{t\in[T]}$ to boost the bandit learning process.
     \end{remark}
 \end{corollary}




\section{Simulation Results}
In this section, we conduct simulation experiments to verify the empirical performance
of proposed algorithms for the tasks of online decoder learning and online codebook learning, respectively. For the former, as previously mentioned, we assume that the transmitter-receiver pair learns  channel decoders $\{\mathbf{G}_t\}_{t\in[T]}$ via a deterministic training sequence of $M$ codewords from a fixed codebook, aimed at improving the transmission performance of data symbols transmitted over time-correlated fading channels.  In the latter case, the transceiver also employs the deterministic training sequence to select optimal  codebooks $\{C_t\}_{t\in [T]}$ from a pre-defined super-codebook $\mathbf{C}$ containing $N$ codebooks for enhancing transmission reliability of data symbols transmitted over the time-correlated additive noise channel. We utilize the average symbol error rate $\frac{1}{t}\sum_{\tau=1}^t \ell_{\tau}(D_{\tau}), t\in[T]$  defined in Eq.~\eqref{eq:fadingloss}
or Eq.~\eqref{eq:correloss} to evaluate the performance of various methods, where $D_{\tau}$ denotes $\mathbf{G}_{\tau}$ or $C_{\tau}$. Building on the coherence time assumption presented in the previous sections,  we can conclude that this metric  reflects the performance of the learned decoder or codebook when applied to random data symbols that contain the information intended for transmission. 

We first introduce the simulation settings for the two tasks. For the online decoder learning task, following the fading channel model outlined in~\cite{DBLP:journals/tsmc/TanSS22}, we model the time-correlated fading channel $Y_t=\mathbf{H}_tX_t+W_t$ as a first-order Markov fading channel: $\mathbf{H}_{t+1}=\sqrt{1-\mu_t}\mathbf{H}_{t}+\sqrt{\mu_t}\boldsymbol{\mathcal{E}}_t$, where $\boldsymbol{\mathcal{E}}_t\in \mathbb{R}^{d\times d}$ is a random matrix and $\mu_t$ is a parameter that governs  the statistical dependency of channel gain between successive  rounds. To ensure that the assumption of bounded expected norm for the channel gain, as stated in Assumption~\ref{assump:env}, is satisfied, we define the parameter $\mu_t$ to decrease with the round $t$. Particularly, we set $\mu_t=\mu^t$ and we choose $\mu$ as $0.96$ in the following simulations.  For the random matrix $\boldsymbol{\mathcal{E}}_t$, we assume that all the elements of $\boldsymbol{\mathcal{E}}_t$ are sampled from a Gaussian mixture  distribution (GMD): $\sum_{k\in[K]}\pi_k\mathcal{N}(\nu_k,\sigma^2_k)$~\cite{10070795} or a  Laplace mixture  distribution (LMD): $\sum_{k\in[K]}\pi_k La(\nu_k,\gamma_k)$~\cite{gai2018speckle}.  We consider that the weighting factor $\{\pi_k\}_{k=1}^K$ is drawn from  a Dirichlet distribution. The mean $\nu_k,\forall k \in [K]$  is drawn from a uniform distribution with the support set $(0,\rho)$, while $\sigma_k$ and $\gamma_k$ is fixed as $1$ for any $k\in[K]$.  The square norm of  $\mathbb{E}[\boldsymbol{\mathcal{E}}_t]$ reflects the degree of cumulative variation $(\Sigma^{\mathbf{H}}_{1:T})^2$, the parameter $\rho$  thus controls the degree of channel variation $(\Sigma^{\mathbf{H}}_{1:T})^2$. As for the online codebook learning task, we adopt the noisy channel model described in~\cite{DBLP:journals/inffus/Caballero-Aguila20}. We  treat the time-correlated additive noise channel $Y_t=X_t+Z_t$ as a first-order Markov additive noise channel: $Z_{t+1}=\sqrt{1-\mu_t}Z_t+\sqrt{\mu_t}\boldsymbol{\epsilon}_t$, where $\boldsymbol{\epsilon}_t$ reprents a random vector and $\mu_t$ also quantifies  the statistical dependency of channel noise across successive rounds.  The random vector $\boldsymbol{\epsilon}_t\in\mathbb{R}^{d}$ is generated by the Gaussian mixture distribution or the Laplace mixture distribution similar to the fading channel scenario. Analogously, we control the degree of channel variation of  $\{Z_t\}_{t\in[T]}$ via $\rho$.

We  first compare the proposed algorithms with various baseline methods introduced below, and  set the code length $d$ as $8$ and the number $M$ of codewords  as $64$ in this experiment. Then we explore the effect of channel variation on the proposed methods, and  set the code length $d$ as $8$ and the number $M$ of codewords  as $16$ for this experiment. The number $K$ of  components in mixture distribution  is set to $3$. 

    \begin{figure*}[ht]
\captionsetup[subfloat]{font=footnotesize}	
\vspace{-0.3in}
\centering
\subfloat[Baselines  for GMD]{\includegraphics[width = 0.25\textwidth]{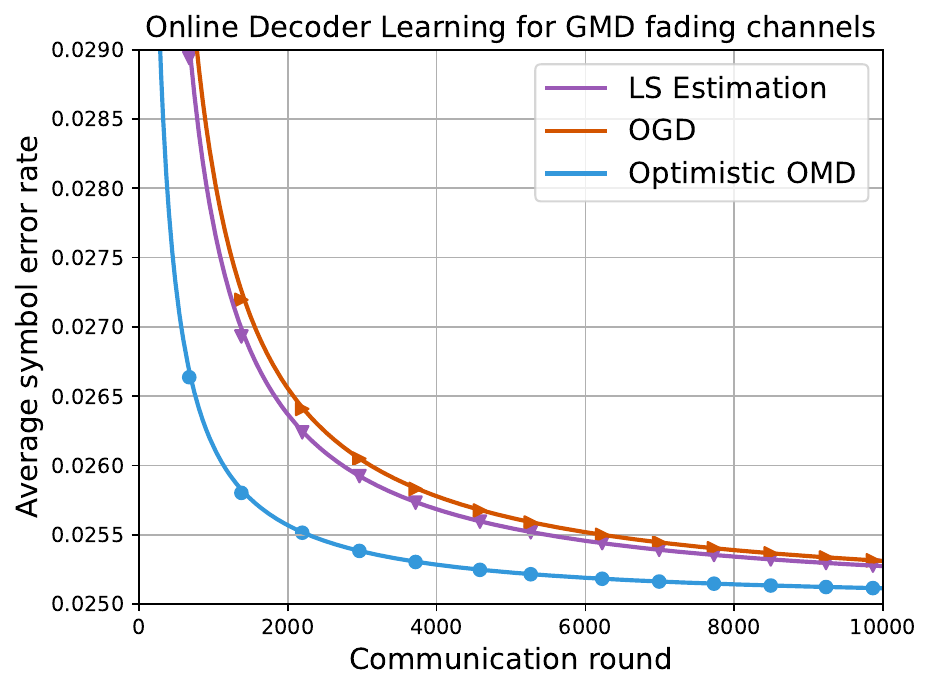}}
   	\subfloat[Baselines for LMD]{\includegraphics[width = 0.25\textwidth]{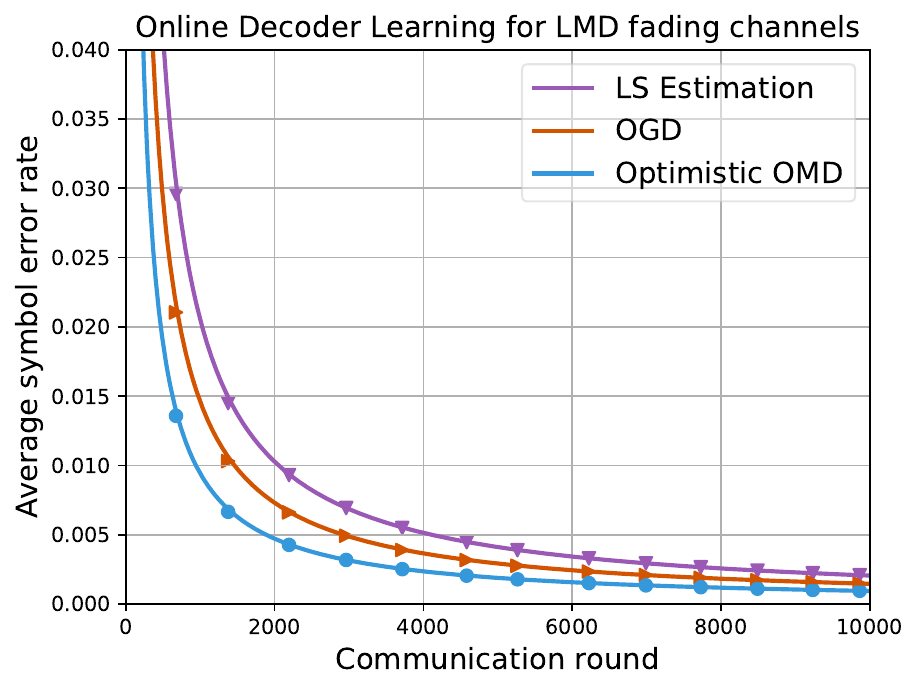}}
	\subfloat[Variation of GMD]{\includegraphics[width = 0.25\textwidth]{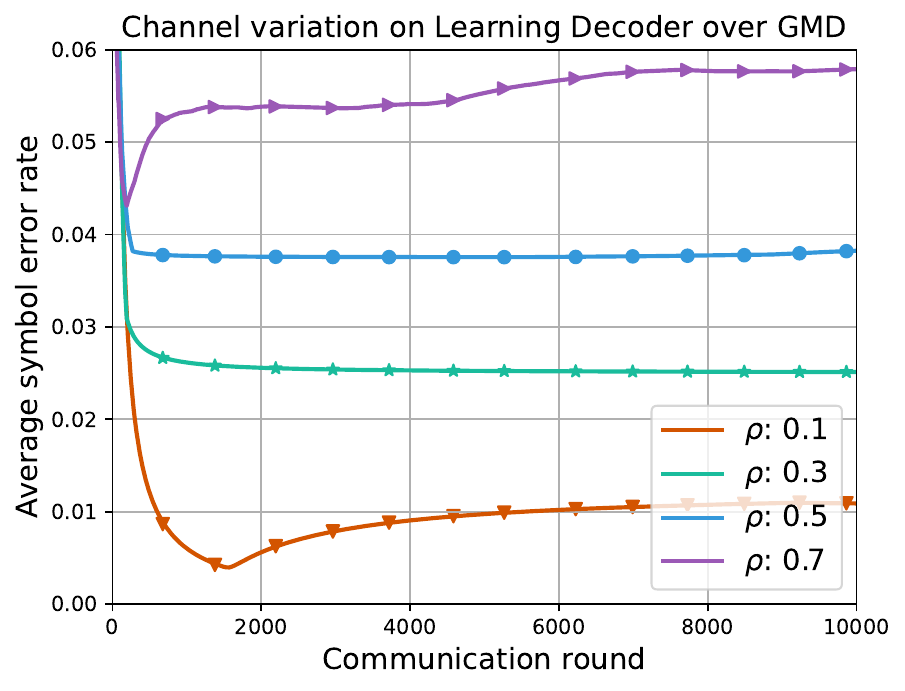}}
 \subfloat[Variation of LMD]{\includegraphics[width = 0.25\textwidth]{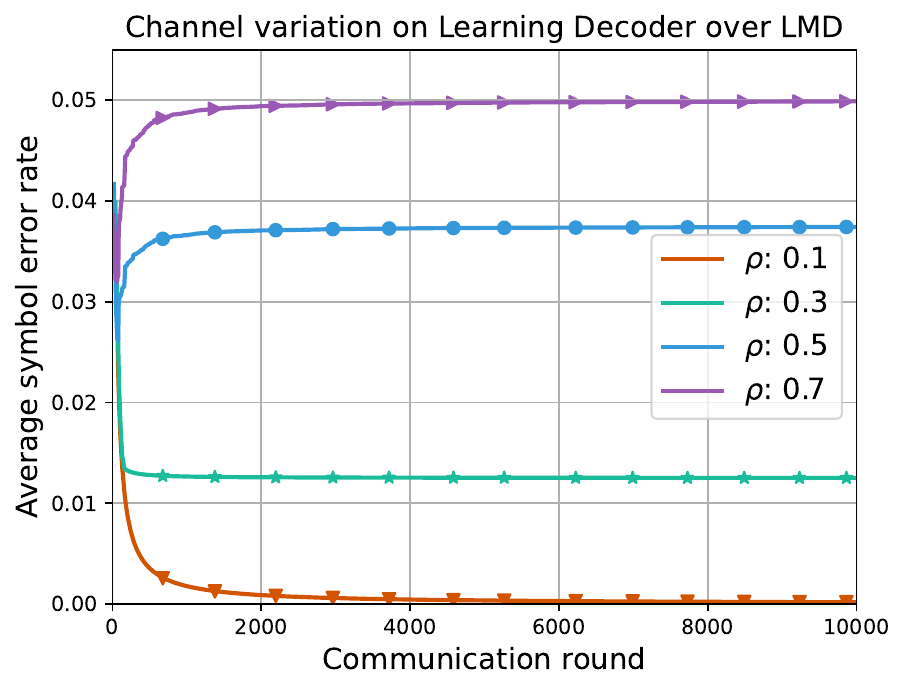}}
\caption{For the online decoder learning task, we compare the proposed Euclidean-regularized optimistic OMD with different baseline methods, and show the effect of  channel variation on the performance of optimistic OMD.}
\label{fig:oco}
\end{figure*}

    \begin{figure*}[ht]
\captionsetup[subfloat]{font=footnotesize}	
\vspace{-0.2in}
\centering
\subfloat[Baselines for GMD]{\includegraphics[width = 0.25\textwidth]{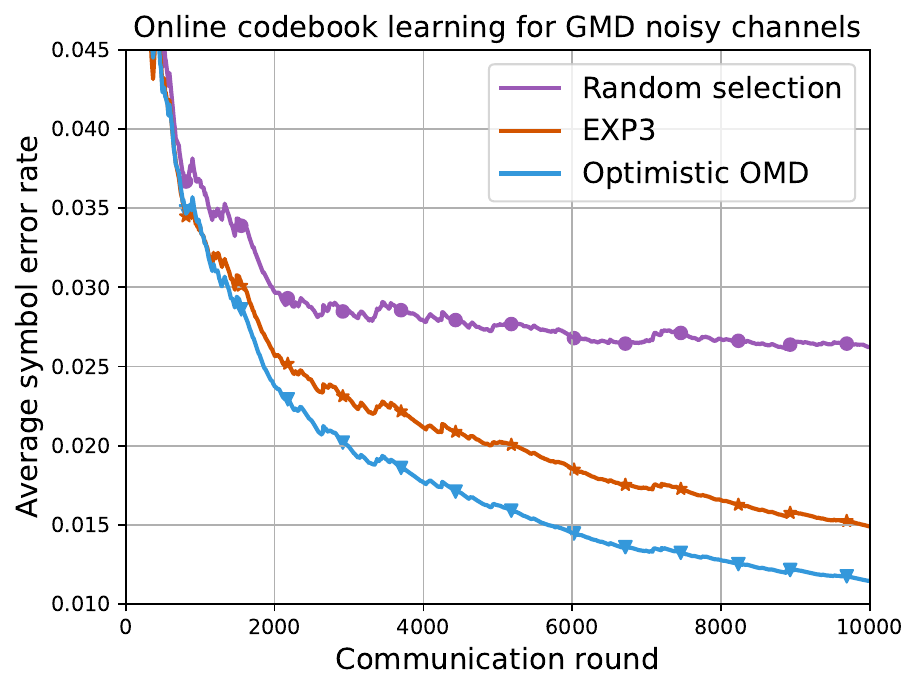}}
   	\subfloat[Baselines for LMD]{\includegraphics[width = 0.25\textwidth]{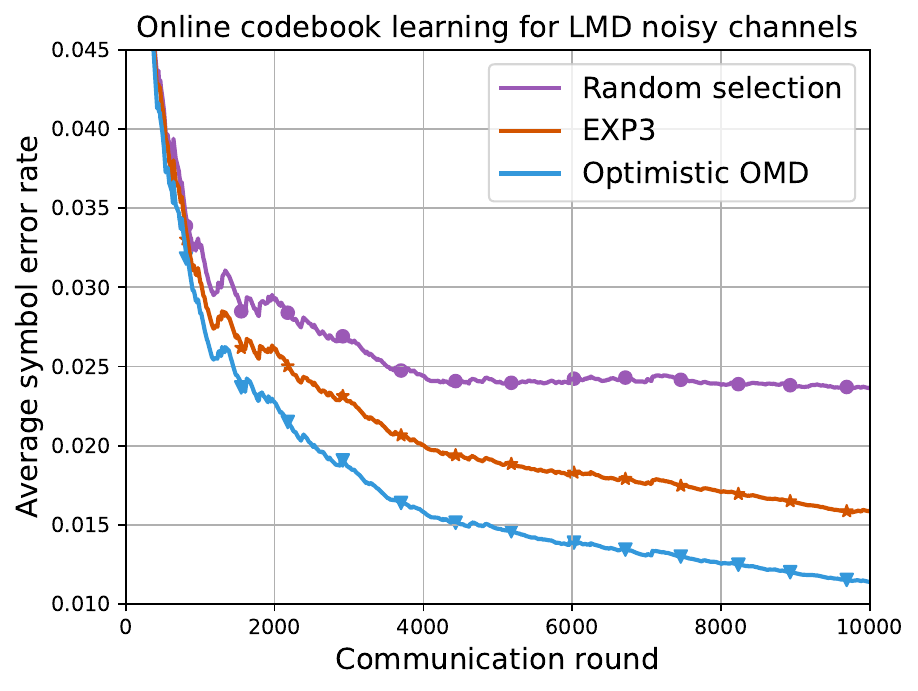}}
	\subfloat[Variation of GMD]{\includegraphics[width = 0.25\textwidth]{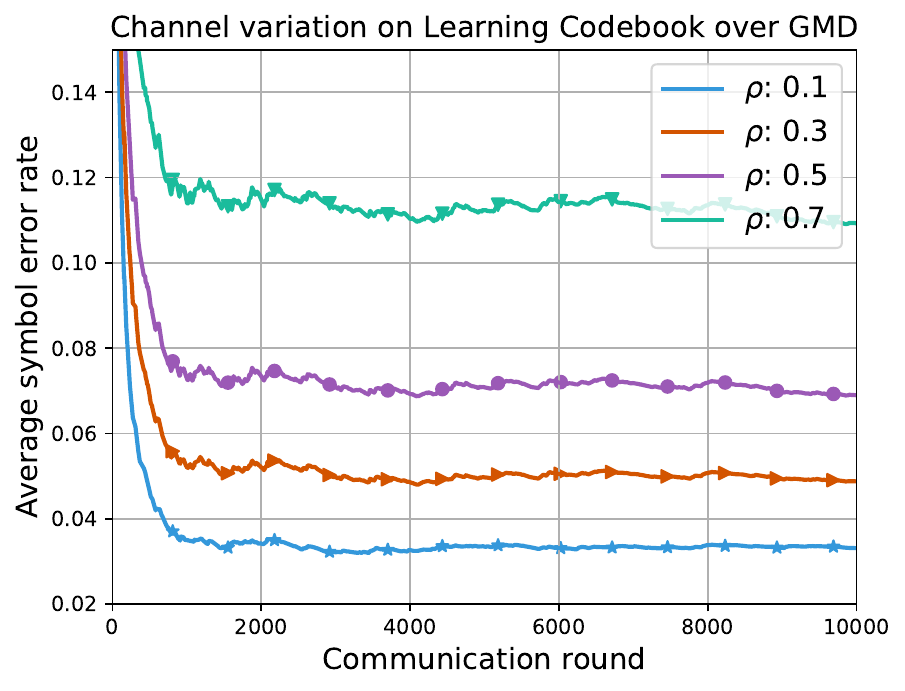}}
 \subfloat[Variation of LMD]{\includegraphics[width = 0.25\textwidth]{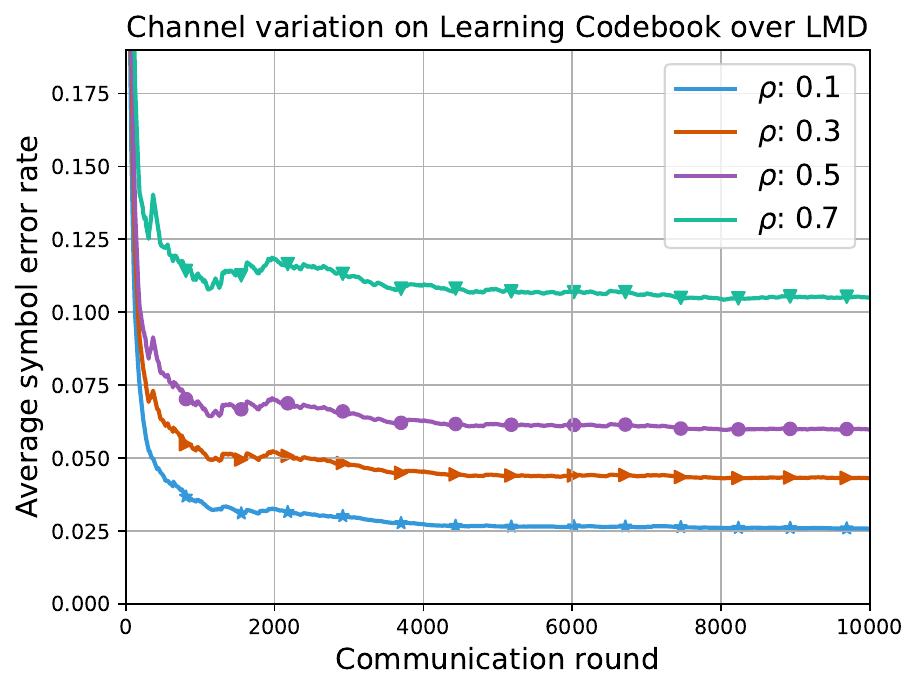}}
\caption{For the online codebook learning task, we compare the proposed Log-barrier-regularized optimistic OMD with different baseline methods, and show the effect of  channel variation on the performance of optimistic OMD.}
 \vspace{-0.2in}
\label{fig:bandit}
\end{figure*}

Initially, we focus on the online decoder learning task.  For each round, we use a fixed codebook $C$ with the constant modulus constraint for transmission, and set the Signal-to-Noise Ratio to $24$dB. In addition, we set the degree $\rho$ of channel variation  to $0.1$ for two channel distributions.
For this task, we compare the proposed method with two baseline methods.  The first is to utilize the least squares estimation~\cite{goldsmith2005wireless} to construct $\mathbf{G}_t$ based on $M$  codewords $\{\mathbf{y}_t^j\}_{j\in[M]}$ received in each round. We choose Online Gradient Descent (OGD) as the second baseline method, which is widely applied in online convex optimization~\cite{shalev2012online,hazan2016introduction}. To investigate the effect of the channel variation on the proposed method, we vary the degree $\rho$ across $\{0.1,0.3,0.5,0.7\}$ for both distributions. As illustrated in Fig.~\ref{fig:oco} (a) and (b), for two channel distributions, the proposed method shows a lower average symbol error rate compared to  other baselines. Additionally, the proposed optimistic OMD method convergences faster than online gradient descent. These results imply that the proposed method leverages distribution dependency to achieve superior performance. Moreover, from the results presented in Fig.~\ref{fig:oco} (c) and (d), the performance of optimistic OMD improves as $\rho$ decreases,  suggesting that the distribution dependency assists the proposed method to learn decoders over time-correlated fading channels, matched with our theoretical findings.

    Next, we concentrate on the online codebook learning task over the time-correlated additive noise channel. The pre-constructed super-codebook $\mathbf{C}$ comprises randomly generated codebooks whose codewords are drawn from a uniform distribution in an element-wise manner. The number $N=|\mathbf{C}|$ of these codebooks is fixed as $100$. For this task, we maintain the degree $\rho$ of  channel variation at $0.01$ for two distributions. Our method is compared with two baseline methods. The first baseline  is to randomly select a codebook from  $\mathbf{C}$ in each round. Then we consider employing a classical multi-armed bandit algorithm known as Exponential-weight for Exploration and Exploitation (EXP3)~\cite{shalev2012online,cesa2006prediction}  as the second baseline. To examine the impact of channel variation on our proposed method, we set the degree $\rho$ of channel variation across $\{0.1,0.3,0.5,0.7\}$ for conducting simulations. All experimental results are showcased in Fig.~\ref{fig:bandit}. In Fig.~\ref{fig:bandit} (a) and (b), the proposed method demonstrates the lowest average symbol error rate over the other baselines across the two distributions. Notably, EXP3 exhibits a worse performance than proposed Log-Barrier regularized optimistic OMD method. It means that EXP3 overlooks utilizing the distribution dependency to boost  bandit learning processes. Besides, as $\rho$ increases, signifying a more pronounced effect of channel variation, our method yields a higher average error rate, indicating that this method in fact leverages the mild environment dynamics to select the optimal codebook.  These empirical observations consistently support and validate our theoretical discoveries.

    \begin{figure*}[ht]
\captionsetup[subfloat]{font=footnotesize}	
\vspace{-0.2in}
\centering
\subfloat[$\mu$ of GMD for decoder learning]{\includegraphics[width =0.25\textwidth]{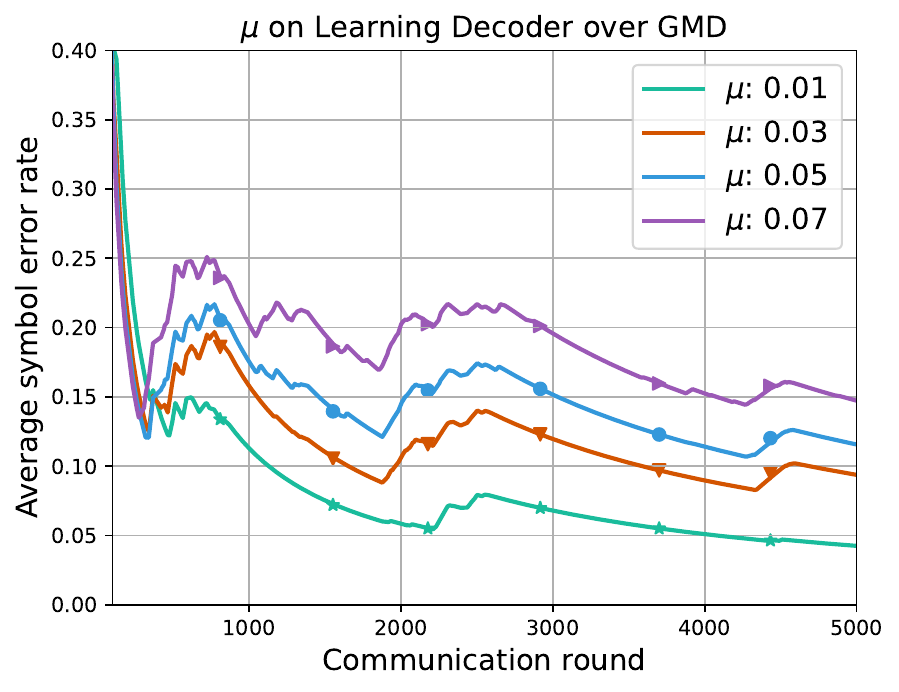}}
\subfloat[$\mu$ of LMD for decoder learning]{\includegraphics[width = 0.25\textwidth]{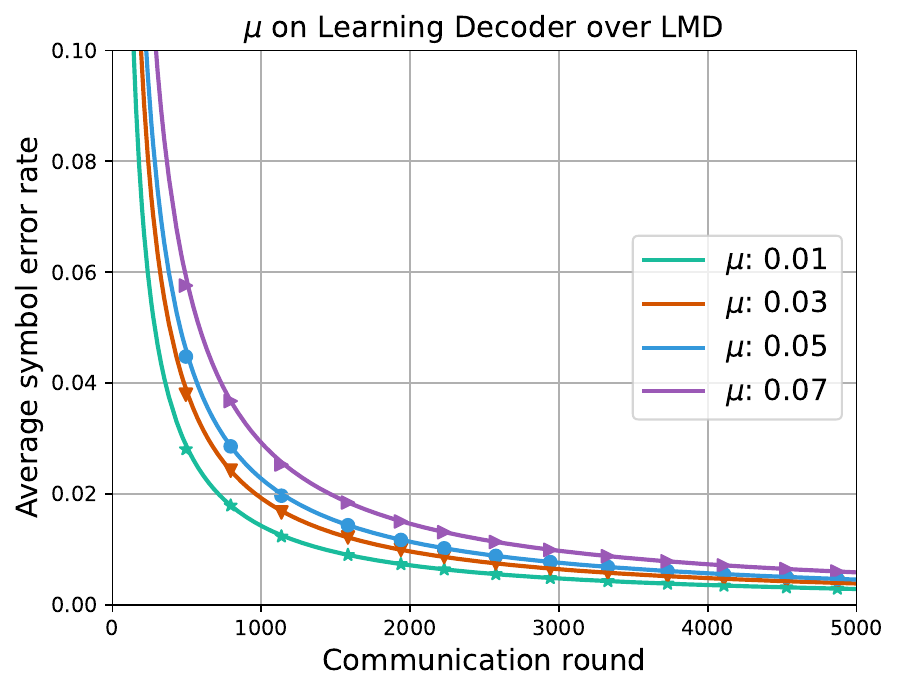}}
	\subfloat[$\mu$ of GMD for codebook learning]{\includegraphics[width = 0.25\textwidth]{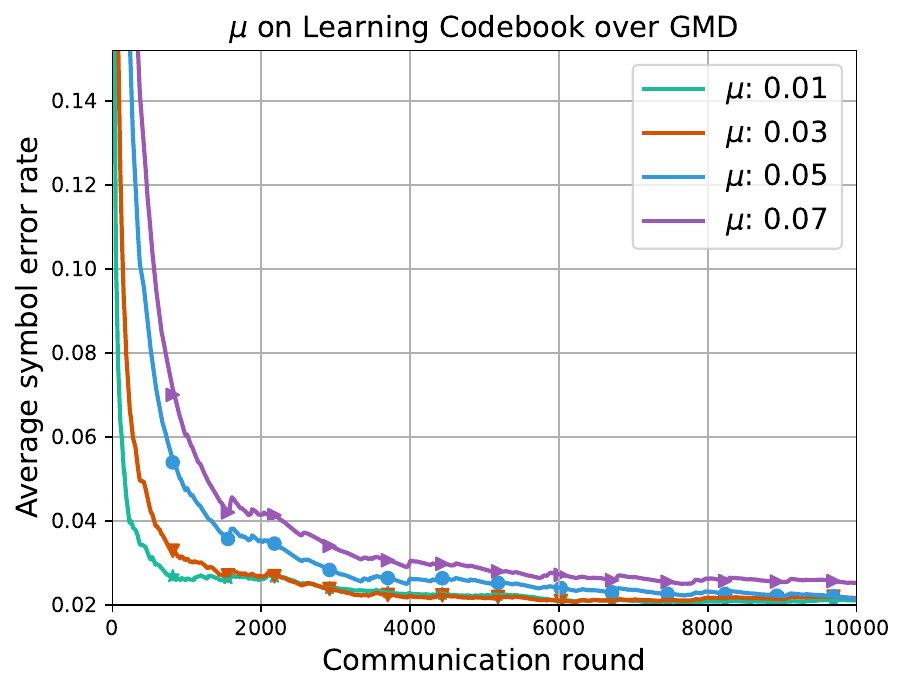}}
 \subfloat[$\mu$ of LMD for codebook learning]{\includegraphics[width =0.25\textwidth]{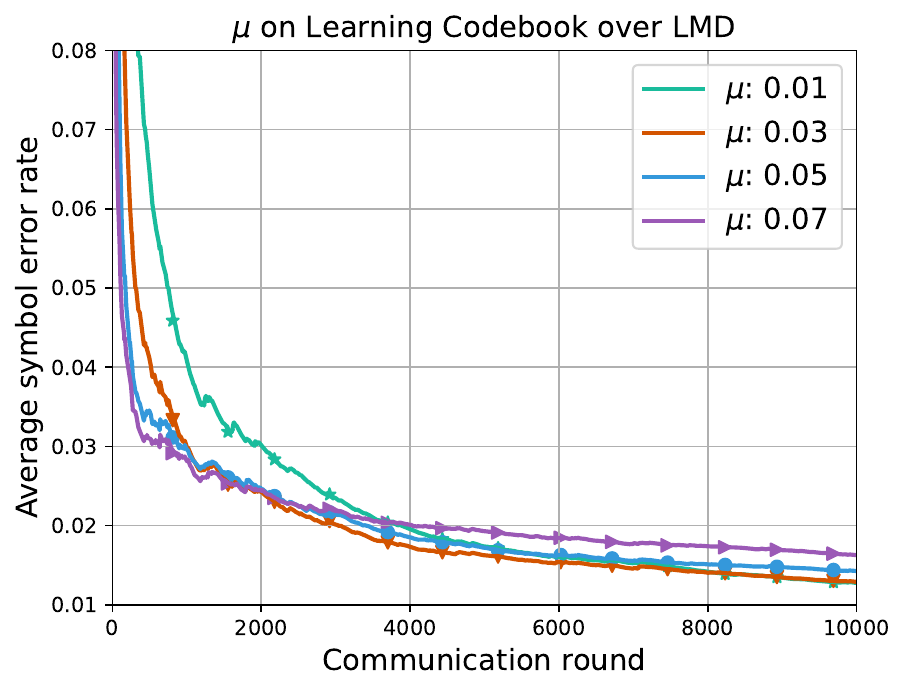}}
\caption{For two considered tasks (online decoder and codebook learning), we explore the effect of different levels of channel correlation on the performance of our proposed optimistic OMD methods by setting different values of $\mu$.}
 \vspace{-0.2in}
\label{fig:mu}
\end{figure*}

Below, we conduct additional simulations to examine our algorithms’ robustness to varying channel correlation levels by setting $\mu_t$ as different constant values $\mu$. We also consider these two primary scenarios: (1) learning decoders for time-correlated fading channels modeled by $\mathbf{H}_{t+1} = \sqrt{1-\mu}\mathbf{H}_t + \sqrt{\mu}\boldsymbol{\mathcal{E}}_t$, where $\boldsymbol{\mathcal{E}}_t$ is generated in the manner described earlier, and (2) learning codebooks for time-correlated additive noise channels governed by $Z_{t+1} = \sqrt{1-\mu}Z_t + \sqrt{\mu}\epsilon_t$, where the simulation setting is similar to that of time-correlated fading channels. By testing multiple correlation levels ($\mu \in \{0.01, 0.03, 0.05, 0.07\}$) under both channels, we obtain a thorough performance assessment across diverse  scenarios.

In Figure~\ref{fig:mu} (a) and (b), we investigate the impact of  $\mu$ on the performance of Euclidean-regularized optimistic OMD in decoder learning.   We find that the average symbol error rate (SER) of optimistic OMD increases with $\mu$, indicating degraded performance as $\mu$ grows from $0.01$ to $0.07$. This is because a larger $\mu$ reduces the temporal correlation of channel gains $\mathbf{H}_t$, making them more statistically independent across rounds. Since optimistic OMD relies on leveraging channel correlation to enhance online learning, its performance deteriorates as $\mu$ increases.   In Figure~\ref{fig:mu} (c) and (d),  we explore the impact of $\mu$ on the performance of Log-barrier-regularized optimistic OMD in codebook learning. The empirical results in Figure~\ref{fig:mu} (c) and (d) are also consistent with our theoretical findings. In Figure~\ref{fig:mu} (d),   although the SER with $\mu = 0.07$ initially decreases faster, it ultimately remains higher than the SER with $\mu = 0.01$.

\begin{figure}[ht]
\centering
\captionsetup[subfloat]{font=footnotesize}	
\subfloat[Baselines for Rayleigh  channels]{\includegraphics[width = 0.25\textwidth]{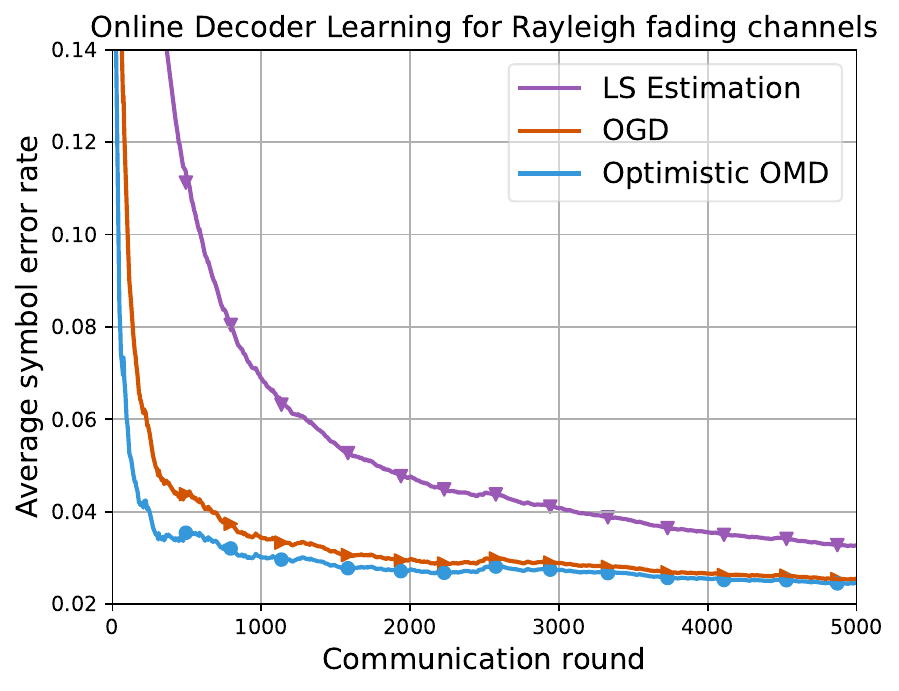}}
   	\subfloat[Baselines for AWGN channels]{\includegraphics[width = 0.25\textwidth]{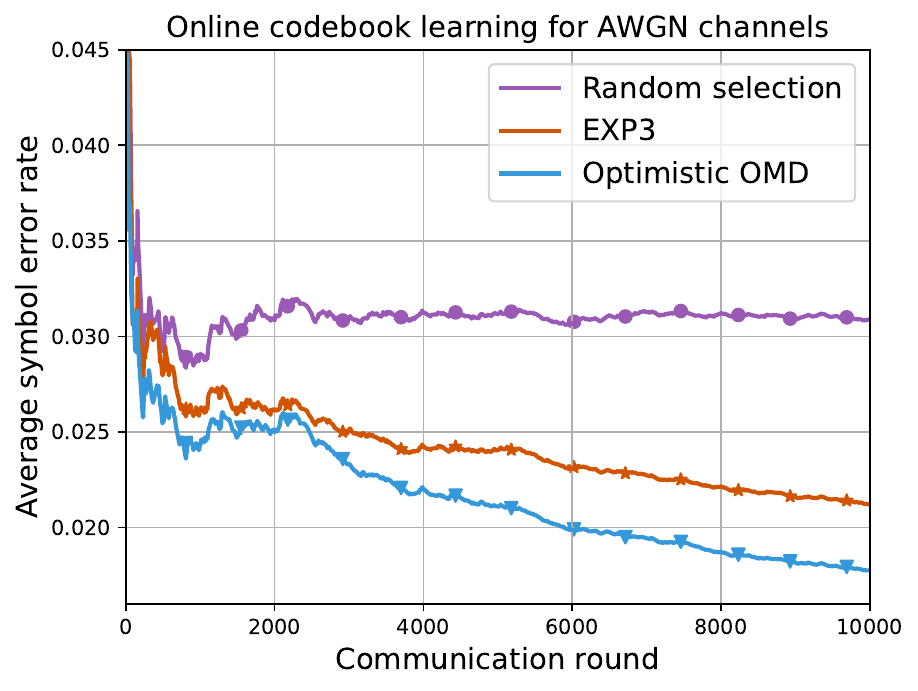}}
\caption{For the online decoder learning task over Rayleigh fading channels, we compare our proposed Euclidean-regularized optimistic OMD algorithm with OGD and LS Estimation. For the online codebook learning task over AWGN channels, we compare our proposed Log-barrier-regularized optimistic OMD algorithm with EXP3 and Random selection.}
\label{fig:common}
\vspace{-0.2in}
\end{figure}

To show the broad applicability of our proposed methods, we then conduct comparative experiments for online decoder and codebook learning in Rayleigh fading channels and AWGN channels, respectively. These two types of channels are commonly used to verify the performance of learning-based communication systems~\cite{bourtsoulatze2019deep}. The channel simulation settings are consistent with those used in most previous studies on learning-based communication systems~\cite{bourtsoulatze2019deep,10418200}. The empirical results are provided in Figure~\ref{fig:common}.

From Figure~\ref{fig:common} (a), we observe that our proposed optimistic OMD method surpasses both OGD and LS Estimation in learning decoders for Rayleigh fading channels. This improvement can be attributed to the additional update step in optimistic OMD, which enhances its performance compared to OGD. Moreover, optimistic OMD significantly outperforms LS Estimation by leveraging historical data more effectively through iterative updates. Figure~\ref{fig:common} (b) further demonstrates that optimistic OMD outperforms baselines such as EXP3 and random selection in learning codebooks for AWGN channels.

\section{Conclusion}

In this paper, we consider the problem of  learning for communication over time-correlated channels via online optimization. To tackle this challenge, we employ the optimistic OMD framework to develop algorithms for designing communication systems. Furthermore, we provide theoretical guarantees for our methods by deriving sub-linear regret bounds on the expected error probability of learned communication schemes. Our theoretical findings confirm that devised algorithms utilize the distribution dependency of time-correlated channels to improve the performance of learned decoders and codebooks. To verify the effectiveness of our approaches, we conduct
extensive simulation experiments and confirm the proposed methods' superiority over other
baselines, thus aligning with our theoretical discoveries. In future work, we will derive lower bounds for the problem of learning to communicate over time-correlated channels and develop novel algorithms capable of leveraging finer-grained channel information.


\appendix

\subsection{Proof of Theorem~\ref{thm:oco}}

\begin{lemma}[Proposition 18 in \cite{DBLP:journals/jmlr/ChiangYLMLJZ12}]
\label{lemma:AAAI23:auxilimary_lemma_1}
    Let $\Omega$ be a convex compact set,
    $\psi$ be a convex function on $\Omega$ and $f'_{t-1}\in\Omega$.
    If
    $
        f^\ast=\mathop{\arg\min}_{f\in\Omega}
        \left\{\langle x,f\rangle+\mathcal{B}_{\psi}(f,f'_{t-1})\right\},
    $
    then, $\forall u\in\Omega$,
    \begin{equation}
        \langle f^\ast-u,x\rangle
        \leq \mathcal{B}_{\psi}(u,f'_{t-1})-\mathcal{B}_{\psi}(u,f^\ast)
        -\mathcal{B}_{\psi}(f^\ast,f'_{t-1}).
    \end{equation}
\end{lemma}

\begin{lemma}[Self-confident tuning~\cite{pogodin2020first}]\label{lemma:selfconfident} Let $\{x_t\}_{t=1}^n$ be  a sequence with $x_t \in [0,B]$ for all $t$. Then
\begin{equation}
    \sum_{t=1}^n \frac{x_t}{\sqrt{1+\sum_{s=1}^{t-1}x_s}}\leq 4\sqrt{1+\sum_{t=1}^n x_t}+B.
\end{equation}
\end{lemma}

Let $\psi_t(\mathbf{G})=\frac{1}{2\eta_t}\Vert \mathbf{G}\Vert^2_F$. 
According to Lemma \ref{lemma:AAAI23:auxilimary_lemma_1}, we can obtain
\begin{equation}
    \begin{aligned}
    &\langle \nabla \Tilde{\ell}_t(\mathbf{G}_t),\mathbf{G}_t-\mathbf{G}\rangle\\
    &=\langle  \mathbf{M}_t,\mathbf{G}_t-\mathbf{G}'_{t+1}\rangle
    +\langle \nabla \Tilde{\ell}_t(\mathbf{G}_t)-\mathbf{M}_t,\mathbf{G}_t-\mathbf{G}'_{t+1}\rangle\\
    &\quad+\langle \nabla \Tilde{\ell}_t(\mathbf{G}_t),\mathbf{G}'_{t+1}-\mathbf{G}\rangle\\
    &\leq \mathcal{B}_{\psi_t}(\mathbf{G}'_{t+1},\mathbf{G}'_t)
    -\mathcal{B}_{\psi_t}(\mathbf{G}'_{t+1},\mathbf{G}_t)-\mathcal{B}_{\psi_t}(\mathbf{G}_t,\mathbf{G}'_t)\\
    &\quad+\langle \nabla \Tilde{\ell}_t(\mathbf{G}_t)-\mathbf{M}_t,\mathbf{G}_t-\mathbf{G}'_{t+1}\rangle\\
    &\quad+\mathcal{B}_{\psi_t}(\mathbf{G},\mathbf{G}'_t)
    -\mathcal{B}_{\psi_t}(\mathbf{G},\mathbf{G}'_{t+1})-\mathcal{B}_{\psi_t}(\mathbf{G}'_{t+1},\mathbf{G}'_t),\\
    &\leq\mathcal{B}_{\psi_t}(\mathbf{G},\mathbf{G}'_t)-\mathcal{B}_{\psi_t}(\mathbf{G},\mathbf{G}'_{t+1})\\
    &\quad+\langle \nabla \Tilde{\ell}_t(\mathbf{G}_t)-\mathbf{M}_t,\mathbf{G}_t-\mathbf{G}'_{t+1}\rangle-\mathcal{B}_{\psi_t}(\mathbf{G}'_{t+1},\mathbf{G}_t)\\
    &=\mathcal{B}_{\psi_t}(\mathbf{G},\mathbf{G}'_t)-\mathcal{B}_{\psi_t}(\mathbf{G},\mathbf{G}'_{t+1})+\frac{\eta_t}{2}\Vert \nabla\Tilde{\ell}_t(\mathbf{G}_t)-\mathbf{M}_t\Vert_F^2\\
    &\quad -\frac{1}{2\eta_t}\Vert \mathbf{G}'_{t+1}-\mathbf{G}_t-\eta_t(\nabla \Tilde{\ell}_t(\mathbf{G}_t)-\mathbf{M}_t)  \Vert_F^2\\
&\leq\mathcal{B}_{\psi_t}(\mathbf{G},\mathbf{G}'_t)-\mathcal{B}_{\psi_t}(\mathbf{G},\mathbf{G}'_{t+1})+\frac{\eta_t}{2}\Vert \nabla\Tilde{\ell}_t(\mathbf{G}_t)-\mathbf{M}_t\Vert_F^2.\\
    \end{aligned}
\end{equation}

 Summing  over $t=1,2,...,T$,
we have
\begin{equation}\label{eq:summary}
    \begin{aligned}
        &\sum_{t=1}^T \langle \nabla \Tilde{\ell}_t(\mathbf{G}_t),\mathbf{G}_t-\mathbf{G}\rangle   \\
        &\leq \underbrace{ \sum_{t=1}^T \frac{1}{2\eta_t}\Big(\Vert \mathbf{G}-\mathbf{G}'_t\Vert_F^2-\Vert \mathbf{G}-\mathbf{G}'_{t+1}\Vert_F^2\Big)}_{\mbox{term (a)}}\\
&\quad+\underbrace{\sum_{t=1}^T\frac{\eta_t}{2}\Vert \nabla\Tilde{\ell}_t(\mathbf{G}_t)-\mathbf{M}_t\Vert_F^2}_{\mbox{term (b)}}.\\
    \end{aligned}
\end{equation}

In the following, we will bound the two terms on the right-hand side respectively.
First, we analyze the term (a). 
Notice that $\eta_t\leq \eta_{t-1}$ and $\mathbf{G}_t\in \mathcal{G}$ satisfies $\Vert \mathbf{G}_t\Vert_F^2\leq D$, so we have
\begin{equation}
    \begin{aligned}
        \mbox{term (a)}&\leq \sum_{t=2}^T \Big(\frac{1}{2\eta_t}-\frac{1}{2\eta_{t-1}}\Big)\Vert \mathbf{G}-\mathbf{G}'_t \Vert_F^2+\frac{1}{2\eta_1}\Vert \mathbf{G}-\mathbf{G}'_1\Vert_F^2\\
        &\leq \frac{D^2}{2\eta_{T}}= D\sqrt{1+\sum_{\tau=1}^{T}\Vert \nabla\Tilde{\ell}_{\tau}(\mathbf{G}_{\tau})-\mathbf{M}_\tau\Vert_F^2}.
    \end{aligned}
\end{equation}

Next, we focus on the term (b). 
Notice that $\mathbf{G}_0$ is a zero matrix,
and $\Vert \nabla\Tilde{\ell}_1(\mathbf{G}_1) \Vert_F^2\leq (2(M-1)d^*L)^2$.
Let $\xi=2(M-1)d^*L$. 
We can bound term (b) as
\begin{equation}
    \begin{aligned}
        \mbox{term (b)}&=\sum_{t=2}^T\frac{\eta_t}{2}(\Vert \nabla\Tilde{\ell}_t(\mathbf{G}_t)-\mathbf{M}_t\Vert_F^2)+\Vert \nabla\Tilde{\ell}_1(\mathbf{G}_1) \Vert_F^2\\
        &\leq D\cdot\sum_{t=2}^T\frac{\Vert \nabla\Tilde{\ell}_t(\mathbf{G}_t)-\mathbf{M}_t\Vert_F^2}{2\sqrt{1+\sum_{\tau=2}^t\Vert \nabla\Tilde{\ell}_t(\mathbf{G}_t)-\mathbf{M}_t\Vert_F^2} }+\xi^2\\
         & \stackrel{\text{(a)}}{\leq}  2D \sqrt{1+\sum_{t=2}^T\Vert \nabla\Tilde{\ell}_t(\mathbf{G}_t)-\mathbf{M}_t\Vert_F^2}
         +(2D+1)\xi^2,\\
    \end{aligned}
\end{equation}
where (a) follows from the Lemma~\ref{lemma:selfconfident}.

Then we substitute the two bounds above into Eq.~\eqref{eq:summary} and we have
\begin{equation}
    \begin{aligned}
        &\sum_{t=1}^T \langle \nabla \Tilde{\ell}_t(\mathbf{G}_t),\mathbf{G}_t-\mathbf{G}\rangle   \\
     &\leq 3D \sqrt{1+\sum_{t=2}^T\Vert \nabla\Tilde{\ell}_t(\mathbf{G}_t)-\mathbf{M}_t\Vert_F^2}+(2D+1)\xi^2.    
     \end{aligned}
\end{equation}

Now we focus on the term $\sum_{t=2}^T\Vert \nabla\Tilde{\ell}_t(\mathbf{G}_t)-\mathbf{M}_t\Vert_F^2$. 
Recall that  $r_t=2d^*DL+\frac{1}{\sqrt{T}}$, thus  the subgradient $\nabla\Tilde{\ell}_t(\mathbf{G}_t)$ defined in Eq.~\eqref{eq:subgradient} becomes
\begin{equation}
\begin{aligned}
\nabla\Tilde{\ell}_t(\mathbf{G}_t)
         &\stackrel{\text{(a)}}{=}\frac{2}{M}\sum_{j=1}^M \sum^M_{j'\neq j} (\mathbf{x}^{j'}-\mathbf{x}^j)(\mathbf{y}_t^j)^T,\\
\end{aligned}
\end{equation}
where (a) holds since  $r_t-(\Vert \mathbf{x}^{j'}-\mathbf{G}_t\mathbf{y}_t^{j}\Vert^2_{2}-\Vert \mathbf{x}^{j}-\mathbf{G}_t\mathbf{y}_t^{j}\Vert^2_{2})=r_t-2 \langle \mathbf{x}^{j}-\mathbf{x}^{j'},\mathbf{G}_t\mathbf{y}_t^j \rangle>0$. The corresponding reason is  presented as follows
 \begin{equation}
    \begin{aligned}
    r_t-2 \langle \mathbf{x}^{j}-\mathbf{x}^{j'},\mathbf{G}_t\mathbf{y}_t^j \rangle&\stackrel{\text{(a)}}{\geq}  r_t-2\Vert \mathbf{x}^{j}-\mathbf{x}^{j'}\Vert_2\Vert \mathbf{G}_t \mathbf{y}_t^j \Vert_2\\
     &\stackrel{\text{(b)}}{\geq}  r_t-2\Vert \mathbf{x}^{j}-\mathbf{x}^{j'}\Vert_2\Vert \mathbf{G}_t\Vert_2\Vert \mathbf{y}_t^j \Vert_2\\
          &\stackrel{\text{(c)}}{\geq}  r_t-2\Vert \mathbf{x}^{j}-\mathbf{x}^{j'}\Vert_2\Vert \mathbf{G}_t\Vert_F\Vert \mathbf{y}_t^j \Vert_2\\
     &\geq r_t-2d^*DL> 0,
    \end{aligned}
\end{equation}
where (a) holds based on the Cauchy-Schwarz inequality, (b) follows from the definition of spectral norm, and (c) follows from the fact that $\Vert \mathbf{G}_t \Vert_2 \leq \Vert \mathbf{G}_t\Vert_F$.

Hence,  $\mathbbm{1}^{j,j'}_t(r_t)$ always equals to $1$ and we have
\begin{equation}
    \begin{aligned}
      \nabla\Tilde{\ell}_t(\mathbf{G}_t)-\mathbf{M}_t  & =\nabla\Tilde{\ell}_t(\mathbf{G}_t)-\nabla\Tilde{\ell}_{t-1}(\mathbf{G}_{t-1})\\
       &=\frac{2}{M}\sum_{j=1}^M \sum^M_{j'\neq j} (\mathbf{x}^{j'}-\mathbf{x}^j)(\mathbf{y}_t^j-\mathbf{y}_{t-1}^j)^T.
    \end{aligned}
\end{equation}

Then we have
\begin{equation}\label{eq:target}
    \begin{aligned}
      &\sum_{t=1}^T \langle \nabla \Tilde{\ell}_t(\mathbf{G}_t),\mathbf{G}_t-\mathbf{G}\rangle \\
      &\leq 3D+(2D+1)\xi^2 +6Dd^*\sqrt{M\sum_{t=2}^T\sum_{j\in[M]}\big\Vert
    \mathbf{y}_t^j-\mathbf{y}_{t-1}^j\big\Vert_2^2}.\\
    \end{aligned}
\end{equation}

Based on this result, we take the expectation of Eq.~\eqref{eq:target} and apply Jensen’s inequality to have
\begin{equation}
    \begin{aligned}
      &\sum_{t=1}^T \mathbb{E}\big[\langle\nabla \Tilde{\ell}_t(\mathbf{G}_t),\mathbf{G}_t-\mathbf{G}\rangle\big] \\
      &\leq 3D+(2D+1)\xi^2 +6Dd^*\sqrt{M\sum_{t=2}^T\sum_{j\in[M]}\mathbb{E}\big\Vert
    \mathbf{y}_t^j-\mathbf{y}_{t-1}^j\big\Vert_2^2}\\
    &\leq\mathcal{O}\Big(\sqrt{M\sum_{t\in [T]}\sum_{j\in [M]}\mathbb{E}\big\Vert
    \mathbf{y}_t^j-\mathbf{y}_{t-1}^j\big\Vert_2^2}\Big).\\
    \end{aligned}
\end{equation}

Based on the theoretical findings in Eq.~\eqref{eq:union}, we have
\begin{equation}
    \begin{aligned}
       &\frac{1}{T} \sum_{t\in[T]}\mathbb{P}_t(\mathbf{G}_t) \\
       &\leq \frac{1}{T}\sum_{t\in[T]}\mathbb{E}[\Tilde{\ell}_t(\mathbf{G}_t)]\\
       &\stackrel{\text{(a)}}{\leq}\frac{1}{T}
       \sum_{t=1}^T \mathbb{E}[\langle \nabla \Tilde{\ell}_t(\mathbf{G}_t),\mathbf{G}_t-\mathbf{G}\rangle]+\frac{1}{T}\sum_{t\in [T]}\mathbb{E}[\Tilde{\ell}_t(\mathbf{G})]\\
        &\leq\mathcal{O}\Big(\sqrt{M\sum_{t\in [T]}\sum_{j\in [M]}\mathbb{E}\big\Vert
    \mathbf{y}_t^j-\mathbf{y}_{t-1}^j\big\Vert_2^2}\Big) +\frac{1}{T}\sum_{t\in [T]}\mathbb{E}[\Tilde{\ell}_t(\mathbf{G})]\\
       &\stackrel{\text{(b)}}{\leq}\mathcal{O}
       \Big(  \frac{M}{T}\sqrt{\sum_{t\in[T]}\big[\gamma_X^2(\gamma_H^2-\mathbb{E}\langle\mathbf{H}_{t},\mathbf{H}_{t-1}\rangle)+\sigma_W^2\big]}\Big)\\
       &\quad+\frac{1}{T}\sum_{t\in [T]}\mathbb{E}[\Tilde{\ell}_t(\mathbf{G})],
    \end{aligned}
\end{equation}
where (a) holds based on the fact that for any $ \mathbf{G} \in \mathcal{G}$,  $   \Tilde{\ell}_t(\mathbf{G}_t)-\Tilde{\ell}_t(G)\leq  \langle \nabla \Tilde{\ell}_t(\mathbf{G}_t),\mathbf{G}_t-\mathbf{G}\rangle$. This fact holds since the surrogate loss $\Tilde{\ell}_t(\mathbf{G}_t)$ is a convex function w.r.t. $\mathbf{G}_t$.  (b) holds based on the fact that $\mathbf{y}_t^j=\mathbf{H}_t\mathbf{x}^j+W_t, \forall j \in [M].$

This completes the proof of Theorem~\ref{thm:oco}.

    \subsection{Proof of Corollary~\ref{corollary:oco}}
Based on the fact that $\mathbf{y}_t^j=\mathbf{H}_t\mathbf{x}^j+W_t$ and $\Vert \mathbf{x}^j \Vert_2 \leq \gamma_X^2$, we find that $\mathbb{E}\Vert \mathbf{y}_t^j-\mathbf{y}_{t-1}^j\Vert_2^2$ satisfies
\begin{equation}
    \begin{aligned}
       &\mathbb{E} \Vert \mathbf{y}_t^j-\mathbf{y}_{t-1}^j\Vert_2^2\\
        &\stackrel{\text{(a)}}{=}  \mathbb{E}\Vert (\mathbf{H}_t-\mathbf{H}_{t-1})\mathbf{x}^j\Vert_2^2+    \mathbb{E}\Vert W_t-W_{t-1}\Vert_2^2\\
        &\leq \gamma^2_X \mathbb{E}\Vert \mathbf{H}_t-\mathbf{H}_{t-1}\Vert_F^2+    \mathbb{E}\Vert W_t-W_{t-1}\Vert_2^2\\
        &=\gamma^2_X \mathbb{E}\Vert \mathbf{H}_t-\mathbf{U}_t+\mathbf{U}_t-\mathbf{U}_{t-1}
        + \mathbf{H}_{t-1}-\mathbf{U}_{t-1}\Vert_F^2\\
        &\quad+    \mathbb{E}\Vert W_t-\mathbb{E}[W]+\mathbb{E}[W]-W_{t-1}\Vert_2^2\\
        &\leq 3\gamma^2_X \Big(\mathbb{E}\Vert  \mathbf{H}_t-\mathbf{U}_t \Vert_F^2+\mathbb{E}\Vert \mathbf{H}_{t-1}-\mathbf{U}_{t-1} \Vert_F^2\\
        &\quad+\mathbb{E}\Vert \mathbf{U}_t-\mathbf{U}_{t-1} \Vert_F^2\Big)+2\sigma_W^2.
    \end{aligned}
\end{equation}
where (a) follows from the fact that the additive white Gaussian noise is zero-mean, and is independent of the channel gain and input.

Based on the proof of Theorem~\ref{thm:oco} and the physical quantities introduced in Section~\ref{sec:symobl}, we have
\begin{equation}
    \begin{aligned}
    &\sum_{t=1}^T \mathbb{E}[\langle \nabla \Tilde{\ell}_t(\mathbf{G}_t),\mathbf{G}_t-\mathbf{G}\rangle ]\\
    &\leq 3D+(2D+1)\xi^2+6Dd^*(M-1)\Big(\gamma_X\sqrt{6(\sigma^{\mathbf{H}}_{1:T})^2}\\
 &\quad+ \gamma_X\sqrt{3(\Sigma^{\mathbf{H}}_{1:T})^2}  +\sqrt{2\sigma_W^2T}\Big) \\
 &\leq \mathcal{O}\Big(M\Big(\sqrt{(\sigma^{\mathbf{H}}_{1:T})^2} +\sqrt{(\Sigma^{\mathbf{H}}_{1:T})^2}+\sqrt{\sigma_W^2T}\Big) \Big).
    \end{aligned}
\end{equation}
This completes the proof of Corollary~\ref{corollary:oco}.

\subsection{Proof of Theorem~\ref{thm:bandit}}
\begin{lemma} [Corollary 9 in \cite{DBLP:conf/colt/WeiL18}]\label{lemma:BroadOMD} For the optimistic OMD  with  the Log-Barrier regularizer, if the hint $\mathbf{m}_t$ satisfies $\mathbf{m}_{t,i}=\boldsymbol{\ell}_{\alpha_i(t),i}$, the loss estimator   $\hat{\boldsymbol{\ell}}$ satisfies $\hat{\boldsymbol{\ell}}_{t,i}=\frac{(\boldsymbol{\ell}_{t,i}-\mathbf{m}_{t,i})\mathbbm{1}\{i_t=i\}}{\mathbf{w}_{t,i}} +\mathbf{m}_{t,i}$, and the learning rate $\eta_t$ satisfies  $\eta_{t,i}=\eta\leq \frac{1}{162}$, using the doubling trick, and we can achieve
      \begin{equation}\label{eq:BROADREG}
          \mathbb{E}[\mbox{Reg}_T]\leq \Tilde{\mathcal{O}}\Big(\sqrt{\sum_{i\in[N]}\sum_{t\in[T]}| \boldsymbol{\ell}_{t,i}-\boldsymbol{\ell}_{t-1,i}|}\Big),
      \end{equation}
      where we take the expectation on the regret over the randomness of algorithm.
    \end{lemma}

We first notice that the function $\sqrt{X}$ is  a concave function in its domain and the Jensen's inequality for concave function is stated as follows: if function $f(X)$ is a concave function w.r.t. the random variable $X$, then we have $\mathbb{E}[f(X)]\leq f(\mathbb{E}[X])$.  Based on the above analysis and Lemma~\ref{lemma:BroadOMD}, taking the expectation on Eq.~\eqref{eq:BROADREG} over the channel noise and applying Jensen’s inequality lead to
    \begin{equation}
        \begin{aligned}
          \mathbb{E}[\mbox{Reg}_T]\leq\Tilde{\mathcal{O}}\Big(\sqrt{\sum_{i\in[N]}\sum_{t\in[T]}\mathbb{E}| \boldsymbol{\ell}_{t,i}-\boldsymbol{\ell}_{t-1,i}|}\Big) .
        \end{aligned}
    \end{equation}

    For the term $  \mathbb{E} \vert \boldsymbol{\ell}_{t,i}-\boldsymbol{\ell}_{t-1,i}\vert$, we can have
   \begin{equation}
        \begin{aligned}
           &\mathbb{E} \vert \boldsymbol{\ell}_{t,i}-\boldsymbol{\ell}_{t-1,i}\vert\\
           &=\mathbb{E}\Big|\frac{1}{M}\sum_{j\in[M]}\big(\mathbbm{1}_{t}^j(i)-\mathbbm{1}_{t-1}^j(i)\big)\Big|\\
            &\leq\frac{1}{M}\sum_{j\in[M]} \mathbb{E}\Big|\mathbbm{1}_{t}^j(i)-\mathbbm{1}_{t-1}^j(i)\Big|\\
            &\leq \frac{1}{M}\sum_{j\in[M]} \big[\mathbb{E}\big|\mathbbm{1}_{t}^j(i)-\mathbb{E}[\mathbbm{1}_{t}^j(i)]\big|+\mathbb{E} \big|\mathbbm{1}_{t-1}^j(i)\\
            &\quad -\mathbb{E}[\mathbbm{1}_{t-1}^j(i)]\big|+\big| \mathbb{E}[\mathbbm{1}_{t}^j(i)]-\mathbb{E}[\mathbbm{1}_{t-1}^j(i)]\big|\big].\\
        \end{aligned}
    \end{equation}

    We then upper-bound the term $\mathbb{E}|\mathbbm{1}_{t}^j(i)-\mathbb{E}[\mathbbm{1}_{t}^j(i)]|$ as
    \begin{equation}
        \begin{aligned}
    \mathbb{E}\big|\mathbbm{1}_{t}^j(i)-\mathbb{E}[\mathbbm{1}_{t}^j(i)]\big|&=     \mathbb{E}\sqrt{\big(\mathbbm{1}_{t}^j(i)-\mathbb{E}[\mathbbm{1}_{t}^j(i)]\big)^2} \\
    &      \stackrel{\text{(a)}}{\leq}
 \sqrt{\mathbb{E}\big(\mathbbm{1}_{t}^j(i)-\mathbb{E}[\mathbbm{1}_{t}^j(i)]\big)^2}=\sigma[\mathbbm{1}_{t}^j(i)],
        \end{aligned}
    \end{equation}
    where (a) holds based on Jensen's inequality and $\sigma[\mathbbm{1}_{t}^j(i)]$ denotes the standard deviation of random variable $\mathbbm{1}_{t}^j(i)$.

    To sum up, we have
    \begin{equation}
   \begin{aligned}
            &\frac{1}{T}\sum_{t\in[T]}\mathbb{P}_t(C_t)- \frac{1}{T}\sum_{t\in[T]}\ell_t(C^*)\\
            &\leq \Tilde{\mathcal{O}}\Big(\frac{1}{T}\sqrt{\sum_{i\in[N]}\sum_{t\in[T]}\frac{1}{M}\sum_{j\in[M]}\big\vert\mathbb{P}^j_{t}(i)-\mathbb{P}^j_{t-1}(i) \big\vert}\\
            &\quad + \frac{1}{T}\sqrt{\sum_{i\in[N]}\sum_{t\in[T]}\sum_{j\in[M]}\frac{\sigma[\mathbbm{1}_t^j(i)]}{M}}\Big),
   \end{aligned}
    \end{equation}
    where $\mathbb{P}_t^j(i)=\mathbb{E}[\mathbbm{1}_t^j(i)]$.

Based on the Pinsker's inequality~\cite{cesa2006prediction}, we can further have $\vert\mathbb{P}^j_{t}(i)-\mathbb{P}^j_{t-1}(i) \vert \leq \sqrt{\frac{1}{2}D_{kl}(\mathbb{P}^j_{t}(i)\Vert \mathbb{P}^j_{t-1}(i))}$ , where  $D_{kl}(P\Vert Q):=\mathbb{E}_P\log\frac{dO}{dQ}$ denotes the relative entropy between probability measures $P$ and $Q$. Then we have
 \begin{equation}
     \begin{aligned}
   & \frac{1}{T}\sum_{t\in[T]}\mathbb{P}_t(C_t)\\
   &\leq     \Tilde{\mathcal{O}}\Big(\frac{1}{T}\sqrt{N\sum_{t\in [T]}\sqrt{D_{kl}(\mathbb{P}_{t}\Vert \mathbb{P}_{t-1})}}\\
       &+ \frac{1}{T}\sqrt{\sum_{i\in[N]}\sum_{t\in[T]}\sum_{j\in[M]}\frac{\sigma[\mathbbm{1}_t^j(i)]}{M}}\Big)  +\frac{1}{T}\sum_{t\in[T]}\ell_t(C^*),\\
     \end{aligned}
 \end{equation}
 where  $D_{kl}(\mathbb{P}_{t}\Vert \mathbb{P}_{t-1}):=\arg\max_{i,j}D_{kl}(\mathbb{P}^j_{t}(i)\Vert \mathbb{P}^j_{t-1}(i))$. 

    This completes the proof of Theorem~\ref{thm:bandit}.

\subsection{Proof of Corollary~\ref{thm:banditGaussian}}
 According to the proof of Theorem~\ref{thm:bandit}, we can directly derive the upper bound of the regret at this scenario where  the number $M$ of codewords satisfies $M=2$ as follows.
\begin{equation}
\begin{aligned}
        \mathbb{E}[\mbox{Reg}_T]&\leq \Tilde{\mathcal{O}}\Big( \sqrt{\sum_{i\in[N]}\sum_{t\in[T]}\sum_{j=1}^2\frac{\sigma[\mathbbm{1}_t^{j}(i)]}{2}}\\
        &\quad + \sqrt{\sum_{i\in[N]}\sum_{t\in[T]}\sum_{j=1}^2\frac{\big\vert\mathbb{P}_t^{j}(i)-\mathbb{P}_{t-1}^{j}(i)\big\vert}{2}}\Big).\\
\end{aligned}
\end{equation}


    Given that $Z_t$ is a zero-mean Gaussian noise with the variance of $\sigma^2_{Z_t}$ at this scenario, we can calculate the corresponding expected error probability $\mathbb{P}_t^{j}(i)=\mathbb{E}[\mathbbm{1}^{j}_t(i)]$ as
\begin{equation}
\begin{aligned}
 \mathbb{P}_t^{j}(i)=\mathbb{E}\big[\mathbbm{1}^{j}_t(i)\big]&=\int_{\frac{d_i}{2}}^{\infty}\frac{1}{\sqrt{2\pi \sigma_{Z_t}^2}}\exp(-\frac{\tau^2}{2\sigma_{Z_t}^2})d\tau.\\
\end{aligned}
\end{equation}

Hence, the term $\big\vert\mathbb{P}_t^{j}(i)-\mathbb{P}_{t-1}^{j}(i)\big\vert$ becomes
\begin{equation}
    \begin{aligned}
        \big|\mathbb{P}_t^{j}(i)-\mathbb{P}_{t-1}^{j}(i)\big|&=\frac{1}{\sqrt{2\pi}}\Big|\int_{\frac{d_i}{2\sigma_{Z_t}}}^{\frac{d_i}{2\sigma_{Z_{t-1}}}}e^{-\frac{\tau^2}{2}}d\tau\Big| \\    &\stackrel{\text{(a)}}{\leq}     \frac{1}{\sqrt{2\pi}} \Big| \frac{d_i}{2\sigma_{Z_t}}-\frac{d_i}{2\sigma_{Z_{t-1}}} \Big|\\
        &=\frac{d_i|\sigma_{Z_t}-\sigma_{Z_{t-1}}|}{2\sqrt{2\pi} \sigma_{Z_t}\sigma_{Z_{t-1}}},
    \end{aligned}
\end{equation}
where  (a) follows from the mean value theorem of integrals and  $|\exp(-x)|\leq 1,\forall x\geq0$.

We then focus on the standard deviation $\sigma[\mathbbm{1}^{j}_t(i)]$ of $\mathbbm{1}^{j}_t(i)$
\begin{equation}
    \begin{aligned}
      \sigma[\mathbbm{1}^{j}_t(i)]
&=\sqrt{\mathbb{E}[\mathbbm{1}^{j}_t(i)]^2-(\mathbb{P}_t^{j}(i))^2}\\
&=\sqrt{\mathbb{P}_t^{j}(i)\Big(1-\mathbb{P}_t^{j}(i)\Big)}\\
&=\sqrt{Q\big(\frac{d_i}{2\sigma_{Z_t}}\big)\Big(1-Q\big(\frac{d_i}{2\sigma_{Z_t}}\big)\Big)}.
    \end{aligned}
\end{equation}

Based on the symmetry of codewords in $2$-ary codebook, we can have
\begin{equation}
\begin{aligned}
        \mathbb{E}[\mbox{Reg}_T]
           &\leq \Tilde{\mathcal{O}}\Big( \sqrt{\sum_{i\in[N]}\sum_{t\in[T]}\sqrt{Q\big(\frac{d_{i}}{2\sigma_{Z_t}}\big)\Big(1-Q\big(\frac{d_{i}}{2\sigma_{Z_t}}\big)\Big)}}\\
           &\quad + \sqrt{\sum_{i\in[N]}\sum_{t\in[T]}|\sigma_{Z_t}-\sigma_{Z_{t-1}}|}\Big).\\
\end{aligned}
\end{equation}
     This completes the proof of Corollary~\ref{thm:banditGaussian}.

\bibliographystyle{ieeetr}
\bibliography{IEEEtran}

\begin{thebibliography}{10}

\bibitem{8542764}
O.~Simeone, ``A very brief introduction to machine learning with applications to communication systems,'' {\em IEEE Transactions on Cognitive Communications and Networking}, vol.~4, no.~4, pp.~648--664, 2018.

\bibitem{bai2019deep}
Q.~Bai, J.~Wang, Y.~Zhang, and J.~Song, ``Deep learning-based channel estimation algorithm over time selective fading channels,'' {\em IEEE Transactions on Cognitive Communications and Networking}, vol.~6, no.~1, pp.~125--134, 2019.

\bibitem{bourtsoulatze2019deep}
E.~Bourtsoulatze, D.~B. Kurka, and D.~G{\"u}nd{\"u}z, ``Deep joint source-channel coding for wireless image transmission,'' {\em IEEE Transactions on Cognitive Communications and Networking}, vol.~5, no.~3, pp.~567--579, 2019.

\bibitem{DBLP:conf/nips/NachmaniW19}
E.~Nachmani and L.~Wolf, ``Hyper-graph-network decoders for block codes,'' in {\em NeurIPS}, pp.~2326--2336, 2019.

\bibitem{DBLP:journals/tccn/OSheaH17}
T.~J. O'Shea and J.~Hoydis, ``An introduction to deep learning for the physical layer,'' {\em {IEEE} Trans. Cogn. Commun. Netw.}, vol.~3, no.~4, pp.~563--575, 2017.

\bibitem{DBLP:journals/wc/HuangGGYZSA20}
H.~Huang, S.~Guo, G.~Gui, Z.~Yang, J.~Zhang, H.~Sari, and F.~Adachi, ``Deep learning for physical-layer {5G} wireless techniques: Opportunities, challenges and solutions,'' {\em {IEEE} Wirel. Commun.}, vol.~27, no.~1, pp.~214--222, 2020.

\bibitem{weinberger2021generalization}
N.~Weinberger, ``Generalization bounds and algorithms for learning to communicate over additive noise channels,'' {\em IEEE Transactions on Information Theory}, vol.~68, no.~3, pp.~1886--1921, 2021.

\bibitem{DBLP:conf/isit/BernardoZE23}
N.~I. Bernardo, J.~Zhu, and J.~S. Evans, ``Learning channel codes from data: Performance guarantees in the finite blocklength regime,'' in {\em {ISIT}}, pp.~2129--2134, {IEEE}, 2023.

\bibitem{liu2024pac}
J.~Liu, W.~Zhang, and H.~V. Poor, ``Pac learnability for reliable communication over discrete memoryless channels,'' {\em arXiv preprint arXiv:2401.13202}, 2024.

\bibitem{shalev2014understanding}
S.~Shalev-Shwartz and S.~Ben-David, {\em Understanding machine learning: From theory to algorithms}.
\newblock Cambridge university press, 2014.

\bibitem{vapnik2013nature}
V.~Vapnik, {\em The nature of statistical learning theory}.
\newblock Springer science \& business media, 2013.

\bibitem{DBLP:journals/inffus/Caballero-Aguila20}
R.~Caballero{-}{\'{A}}guila, A.~Hermoso{-}Carazo, and J.~Linares{-}P{\'{e}}rez, ``Networked fusion estimation with multiple uncertainties and time-correlated channel noise,'' {\em Inf. Fusion}, vol.~54, pp.~161--171, 2020.

\bibitem{DBLP:journals/tmc/YaoBS23}
G.~Yao, A.~M. Bedewy, and N.~B. Shroff, ``Age-optimal low-power status update over time-correlated fading channel,'' {\em {IEEE} Trans. Mob. Comput.}, vol.~22, no.~8, pp.~4500--4514, 2023.

\bibitem{DBLP:journals/tsmc/TanSS22}
H.~Tan, B.~Shen, and H.~Shu, ``Robust recursive filtering for stochastic systems with time-correlated fading channels,'' {\em {IEEE} Trans. Syst. Man Cybern. Syst.}, vol.~52, no.~5, pp.~3102--3112, 2022.

\bibitem{DBLP:journals/access/ZhuYCTFWL18}
Q.~Zhu, Y.~Yang, X.~Chen, Y.~Tan, Y.~Fu, C.~Wang, and W.~Li, ``A novel 3d non-stationary vehicle-to-vehicle channel model and its spatial-temporal correlation properties,'' {\em {IEEE} Access}, vol.~6, pp.~43633--43643, 2018.

\bibitem{DBLP:journals/isci/WangD24}
J.~Wang and J.~Dong, ``Asynchronous dissipative control for nonhomogeneous {Markov} jump systems with dual {Markovian} wireless fading channels,'' {\em Inf. Sci.}, vol.~659, p.~120071, 2024.

\bibitem{DBLP:journals/tit/LarranagaADP18}
M.~Larra{\~{n}}aga, M.~Assaad, A.~Destounis, and G.~S. Paschos, ``Asymptotically optimal pilot allocation over markovian fading channels,'' {\em {IEEE} Trans. Inf. Theory}, vol.~64, no.~7, pp.~5395--5418, 2018.

\bibitem{DBLP:journals/tcom/ChavaliS13}
V.~G. Chavali and C.~R. C.~M. da~Silva, ``Classification of digital amplitude-phase modulated signals in time-correlated non-gaussian channels,'' {\em {IEEE} Trans. Commun.}, vol.~61, no.~6, pp.~2408--2419, 2013.

\bibitem{542437}
B.~Sadler, ``Detection in correlated impulsive noise using fourth-order cumulants,'' {\em IEEE Transactions on Signal Processing}, vol.~44, no.~11, pp.~2793--2800, 1996.

\bibitem{mohri2018foundations}
M.~Mohri, A.~Rostamizadeh, and A.~Talwalkar, {\em Foundations of machine learning}.
\newblock MIT press, 2018.

\bibitem{shalev2012online}
S.~Shalev-Shwartz {\em et~al.}, ``Online learning and online convex optimization,'' {\em Foundations and Trends{\textregistered} in Machine Learning}, vol.~4, no.~2, pp.~107--194, 2012.

\bibitem{hazan2016introduction}
E.~Hazan {\em et~al.}, ``Introduction to online convex optimization,'' {\em Foundations and Trends{\textregistered} in Optimization}, vol.~2, no.~3-4, pp.~157--325, 2016.

\bibitem{DBLP:journals/spl/WuXYL24}
Z.~Wu, Z.~Xu, H.~Yu, and J.~Liu, ``Information-theoretic generalization analysis for topology-aware heterogeneous federated edge learning over noisy channels,'' {\em {IEEE} Signal Process. Lett.}, vol.~31, pp.~691--695, 2024.

\bibitem{DBLP:journals/tvt/WuXZLL24}
Z.~Wu, Z.~Xu, D.~Zeng, J.~Li, and J.~Liu, ``Topology learning for heterogeneous decentralized federated learning over unreliable {D2D} networks,'' {\em {IEEE} Trans. Veh. Technol.}, vol.~73, no.~8, pp.~12201--12206, 2024.

\bibitem{DBLP:journals/tcom/ZhangHZY20}
J.~Zhang, Y.~Huang, Y.~Zhou, and X.~You, ``Beam alignment and tracking for millimeter wave communications via bandit learning,'' {\em {IEEE} Trans. Commun.}, vol.~68, no.~9, pp.~5519--5533, 2020.

\bibitem{wei2022fast}
Y.~Wei, Z.~Zhong, and V.~Y. Tan, ``Fast beam alignment via pure exploration in multi-armed bandits,'' {\em IEEE Transactions on Wireless Communications}, vol.~22, no.~5, pp.~3264--3279, 2022.

\bibitem{DBLP:journals/jmlr/ChiangYLMLJZ12}
C.~Chiang, T.~Yang, C.~Lee, M.~Mahdavi, C.~Lu, R.~Jin, and S.~Zhu, ``Online optimization with gradual variations,'' in {\em {COLT}}, vol.~23 of {\em {JMLR} Proceedings}, pp.~6.1--6.20, JMLR.org, 2012.

\bibitem{DBLP:conf/colt/WeiL18}
C.~Wei and H.~Luo, ``More adaptive algorithms for adversarial bandits,'' in {\em {COLT}}, vol.~75 of {\em Proceedings of Machine Learning Research}, pp.~1263--1291, {PMLR}, 2018.

\bibitem{DBLP:conf/colt/RakhlinS13}
A.~Rakhlin and K.~Sridharan, ``Online learning with predictable sequences,'' in {\em {COLT}}, vol.~30 of {\em {JMLR} Workshop and Conference Proceedings}, pp.~993--1019, JMLR.org, 2013.

\bibitem{DBLP:journals/tccn/CaciularuB20}
A.~Caciularu and D.~Burshtein, ``Unsupervised linear and nonlinear channel equalization and decoding using variational autoencoders,'' {\em {IEEE} Trans. Cogn. Commun. Netw.}, vol.~6, no.~3, pp.~1003--1018, 2020.

\bibitem{10423294}
K.~Weththasinghe, B.~Jayawickrama, and Y.~He, ``Machine learning-based channel estimation for 5g new radio,'' {\em IEEE Wireless Communications Letters}, vol.~13, no.~4, pp.~1133--1137, 2024.

\bibitem{8919799}
F.~Carpi, C.~Häger, M.~Martalò, R.~Raheli, and H.~D. Pfister, ``Reinforcement learning for channel coding: Learned bit-flipping decoding,'' in {\em 2019 57th Annual Allerton Conference on Communication, Control, and Computing (Allerton)}, pp.~922--929, 2019.

\bibitem{DBLP:journals/jstsp/NachmaniMLGBB18}
E.~Nachmani, E.~Marciano, L.~Lugosch, W.~J. Gross, D.~Burshtein, and Y.~Be'ery, ``Deep learning methods for improved decoding of linear codes,'' {\em {IEEE} J. Sel. Top. Signal Process.}, vol.~12, no.~1, pp.~119--131, 2018.

\bibitem{DBLP:journals/twc/ShlezingerFEG20}
N.~Shlezinger, N.~Farsad, Y.~C. Eldar, and A.~J. Goldsmith, ``Viterbinet: {A} deep learning based viterbi algorithm for symbol detection,'' {\em {IEEE} Trans. Wirel. Commun.}, vol.~19, no.~5, pp.~3319--3331, 2020.

\bibitem{8259241}
F.~Liang, C.~Shen, and F.~Wu, ``An iterative bp-cnn architecture for channel decoding,'' {\em IEEE Journal of Selected Topics in Signal Processing}, vol.~12, no.~1, pp.~144--159, 2018.

\bibitem{adiga2024generalization}
S.~Adiga, X.~Xiao, R.~Tandon, B.~Vasi{\'c}, and T.~Bose, ``Generalization bounds for neural belief propagation decoders,'' {\em IEEE Transactions on Information Theory}, 2024.

\bibitem{DBLP:journals/tit/TsvieliW23}
A.~Tsvieli and N.~Weinberger, ``Learning maximum margin channel decoders,'' {\em {IEEE} Trans. Inf. Theory}, vol.~69, no.~6, pp.~3597--3626, 2023.

\bibitem{cesa2006prediction}
N.~Cesa-Bianchi and G.~Lugosi, {\em Prediction, learning, and games}.
\newblock Cambridge university press, 2006.

\bibitem{10772999}
Z.~Xu, Z.~Zhang, S.~Wang, X.~Hu, Y.~Jia, and B.~Ren, ``Energy-constrained distributed mac in cr-iot networks: A budgeted multi-player multi-armed bandit approach,'' {\em IEEE Transactions on Cognitive Communications and Networking}, pp.~1--1, 2024.

\bibitem{DBLP:journals/corr/abs-2404-09494}
J.~Li, Z.~Xu, Z.~Wu, and I.~King, ``On the necessity of collaboration in online model selection with decentralized data,'' {\em CoRR}, vol.~abs/2404.09494, 2024.

\bibitem{DBLP:conf/icml/ChenT0023}
S.~Chen, W.~Tu, P.~Zhao, and L.~Zhang, ``Optimistic online mirror descent for bridging stochastic and adversarial online convex optimization,'' in {\em {ICML}}, vol.~202 of {\em Proceedings of Machine Learning Research}, pp.~5002--5035, {PMLR}, 2023.

\bibitem{goldsmith2005wireless}
A.~Goldsmith, {\em Wireless communications}.
\newblock Cambridge university press, 2005.

\bibitem{9425522}
Y.~Heng and J.~G. Andrews, ``Machine learning-assisted beam alignment for mmwave systems,'' {\em IEEE Transactions on Cognitive Communications and Networking}, vol.~7, no.~4, pp.~1142--1155, 2021.

\bibitem{he2022qcqp}
X.~He and J.~Wang, ``Qcqp with extra constant modulus constraints: Theory and application to sinr constrained mmwave hybrid beamforming,'' {\em IEEE Transactions on Signal Processing}, vol.~70, pp.~5237--5250, 2022.

\bibitem{Auer2002Adaptive}
P.~Auer, N.~Cesa{-}Bianchi, and C.~Gentile, ``Adaptive and {Self-Confident} {On-Line} {Learning Algorithms},'' {\em Journal of Computer and System Sciences}, vol.~64, no.~1, pp.~48--75, 2002.

\bibitem{DBLP:journals/jsac/ForneyW89}
G.~D.~F. Jr. and L.~Wei, ``Multidimensional constellations. {I. Introduction}, figures of merit, and generalized cross constellations,'' {\em {IEEE} J. Sel. Areas Commun.}, vol.~7, no.~6, pp.~877--892, 1989.

\bibitem{zamir2014lattice}
R.~Zamir, {\em Lattice Coding for Signals and Networks: A Structured Coding Approach to Quantization, Modulation, and Multiuser Information Theory}.
\newblock Cambridge University Press, 2014.

\bibitem{10070795}
X.~Chen, H.~Wu, Y.~Cheng, W.~Feng, and Y.~Guo, ``Joint design of transmit sequence and receive filter based on riemannian manifold of gaussian mixture distribution for mimo radar,'' {\em IEEE Transactions on Geoscience and Remote Sensing}, vol.~61, pp.~1--13, 2023.

\bibitem{gai2018speckle}
S.~Gai, B.~Zhang, C.~Yang, and L.~Yu, ``Speckle noise reduction in medical ultrasound image using monogenic wavelet and laplace mixture distribution,'' {\em Digital Signal Processing}, vol.~72, pp.~192--207, 2018.

\bibitem{10418200}
S.~Adiga, X.~Xiao, R.~Tandon, B.~Vasić, and T.~Bose, ``Generalization bounds for neural belief propagation decoders,'' {\em IEEE Transactions on Information Theory}, vol.~70, no.~6, pp.~4280--4296, 2024.

\bibitem{pogodin2020first}
R.~Pogodin and T.~Lattimore, ``On first-order bounds, variance and gap-dependent bounds for adversarial bandits,'' in {\em Uncertainty in Artificial Intelligence}, pp.~894--904, PMLR, 2020.

\end{thebibliography}
\end{document}